\documentclass[preprint,12pt]{elsarticle}
\usepackage{cmap}
\usepackage[T1]{fontenc}
\usepackage{lmodern}
\usepackage[utf8]{inputenc}
\usepackage{amsmath,amssymb}
\usepackage{graphicx}
\usepackage[table]{xcolor}
\usepackage{booktabs}
\definecolor{noteA}{RGB}{214,231,245}
\definecolor{noteB}{RGB}{222,240,214}
\definecolor{noteC}{RGB}{253,243,208}
\definecolor{noteD}{RGB}{233,222,243}
\usepackage{array}
\usepackage{longtable}
\usepackage{url}
\usepackage[hidelinks]{hyperref}
\usepackage{xurl}
\journal{Journal of Parallel and Distributed Computing}
\begin{document}
\begin{frontmatter}
\title{A Feature-Major Codebook for Memory-Efficient Sparse-Binary Self-Organizing Maps: Scaling a MEDLINE Atlas to 1.05 Million Neurons on a Single Consumer GPU}
\author{Andrew J. Amos}
\ead{andrew.amos@my.jcu.edu.au}
\ead[url]{https://orcid.org/0000-0002-9145-0212}
\affiliation{organization={College of Medicine and Dentistry, James Cook University},
             city={Townsville}, state={Queensland}, country={Australia}}

\begin{abstract}
A self-organising map turns a large corpus into a browsable two-dimensional atlas, but building one at MEDLINE scale has been impractical: the best-matching-unit (BMU) search that dominates training is bound by the bandwidth needed to read the codebook every epoch. I show that this bottleneck is largely an artefact of codebook layout. Storing it feature-major with each feature's weights contiguous, W[v$\cdot$M+i], recasts the search as a tiled sparse-dense product in which every loaded weight column is reused across a tile of samples. Varying only the layout, with implementation, precision and update rule held fixed, accelerates the BMU search by 4.5-8.5$\times$. Because an exact-argmin BMU is invariant to how the codebook is stored, this gain costs nothing: held-out quantisation error agrees with a cuSPARSE baseline to within 0.5\% at every map size. Against that baseline the advantage is a crossover rather than a constant: cuSPARSE.SOM is faster at small maps, SparseBin.SOM is 1.5$\times$ faster at 128$\times$128 and 2.6$\times$ at 256$\times$256, and at 512$\times$512 it is the only one that runs on 24 GB without re-engineering its memory path. Paired with a radius-independent box-blur update and a convergence-based stopping rule, it trains a converged map over 29.9 million MEDLINE articles in about 72 s at 64$\times$64 on one 24 GB GPU, and accommodates 262,144 neurons (512$\times$512 edges) where every alternative algorithm I tested exceeds memory constraints. On a 141 GB H200 it reaches 1,048,576 neurons (1024$\times$1024 edges) - to my knowledge the largest self-organising map yet reported. Held-out error follows a smooth power law with no elbow across three decades of map size, so the limit on resolution is compute rather than any breakpoint in the data. A post-submission addendum, in which both implementations were tuned symmetrically and retrained, reports the search accelerated a further 5.6-10.1$\times$, bringing that 64$\times$64 run to about 13 s. It also removes the crossover, SparseBin.SOM being faster at every map size against a tuned baseline, raises the margins at 128$\times$128 to roughly 385$\times$ MedSOM and 3,000$\times$ somoclu, and narrows two mechanism claims made here. At matched work, in the configuration benchmarked here, the design is $\sim$82$\times$ faster than MedSOM, the CUDA implementation behind our earlier MEDLINE atlases and, at 128$\times$128, 621$\times$ faster than the best available multicore-CPU library.
\end{abstract}
\begin{keyword}
self-organizing maps \sep GPU computing \sep sparse matrix kernels \sep memory-bandwidth optimisation \sep CUDA \sep MEDLINE/MeSH
\end{keyword}
\end{frontmatter}

\section{Introduction}
The organisation of medical knowledge has practical consequences. Decisions about what to teach, what to research, and what to prioritise still rest largely on expert judgement which, however well-informed, carries the systematic biases of its authors - with documented costs to the diversity of the medical workforce and the completeness of medical curricula \cite{amos2021}. The peer-reviewed literature indexed in MEDLINE is the most comprehensive and least biased record of medical knowledge available, and techniques that summarise it can offer objective, structural evidence to complement expert judgement in curriculum development, for example by surfacing emerging topics for curriculum renewal \cite{amos2022}. Among such techniques the self-organising map (SOM) is distinctive in producing a discrete, addressable, two-dimensional atlas in which topical neighbourhoods are spatially contiguous and which can be read much as one reads a geographic map \cite{amos2024a,amos2024b}. Realising that promise across the whole corpus is, however, first of all a computational problem - and it is that problem this paper addresses.

\sloppy A self-organizing map projects high-dimensional data onto a low-dimensional lattice of prototype vectors (called neurons in this manuscript) such that nearby neurons respond to similar inputs. The result is a discrete, addressable, and browsable atlas: every input is assigned to a grid cell, and the grid itself is laid out so that topical neighbourhoods are spatially contiguous. For a corpus on the scale of MEDLINE - tens of millions of articles indexed against a controlled vocabulary of tens of thousands of Medical Subject Headings (MeSH) \cite{lipscomb2000} - such an atlas is an attractive organising structure, but producing it is computationally demanding.

The cost of SOM training is dominated by the best-matching-unit (BMU) search: for each input, the algorithm must find the nearest of M prototype vectors (representing neurons). At corpus scale this search reads the codebook - the M$\times$V table of prototype weights - repeatedly across many epochs. When the codebook is laid out node-major (each neuron's V weights stored contiguously), the per-warp access pattern is a strided gather across non-contiguous memory; at a codebook size far exceeding the L2 cache, every gather becomes an HBM round-trip and the kernel is bound by uncoalesced memory traffic. The BMU search is, in short, memory-bandwidth-bound, and the lever for accelerating it is the codebook layout.

In this paper I demonstrate that storing the codebook feature-major - the M weights for a given feature stored contiguously, W[v$\cdot$M+i] - is the more efficient layout for sparse inputs. It makes each feature's weight column a coalesced read, and it enables a tiled BMU kernel in which a loaded weight column is reused across all samples in a tile that share that feature. The material difference is the access pattern rather than the values stored, so it does not depend on the inputs being binary; the same layout drives a sparse-float variant of the design (Table~\ref{tab:6}). Binary inputs are a specialisation on top of it, worth a further 2.1-2.8$\times$ speedup in per-epoch training time (5.5). I pair this layout with a factorised box-blur neighbourhood update whose cost is independent of the neighbourhood radius, and a convergence-based stopping rule, to obtain an end-to-end system that trains over a MEDLINE-scale corpus on a single commodity GPU.

A central observation organises the evaluation. An exact-argmin BMU returns the same assignment regardless of how the codebook is stored or which kernel computes it, provided the sample tiling is held fixed; the layout therefore affects only speed, never quality. This lets me make the efficiency claim rigorously: I demonstrate matched clustering quality (quantisation error, topographic error, dead-unit fraction) between SparseBin.SOM and a strong same-GPU library baseline, and attribute the entire speed-up to the BMU/layout axis. I deliberately restrict `quality' here to intrinsic SOM quality; the scientific validity of the resulting MEDLINE atlas - whether it recovers the MeSH and citation-derived knowledge structure - is the subject of a companion paper \cite{amosprep}. The present method also scales our prior MEDLINE-SOM programme - a 350$\times$350 cartographic atlas of medical knowledge \cite{amos2024a}, externally validated against an expert-derived textbook knowledge structure \cite{amos2024b} - by more than an order of magnitude in neuron count.

Contributions. This paper makes the following contributions:

\begin{itemize}\item A feature-major codebook layout for sparse SOMs, with the memory-access argument for why it coalesces reads and enables tile-level weight reuse (3.2).\end{itemize}
\begin{itemize}\item An SpMM-tile BMU kernel that reads only the distinct feature columns a tile touches, reuses each across the tile's samples, and fuses the argmin into the node sweep (3.3).\end{itemize}
\begin{itemize}\item Empirical confirmation that the BMU/layout axis affects only speed, isolating quality to the update rule, which makes the speed-up a verifiable free lunch (3.4, 5.3).\end{itemize}
\begin{itemize}\item An end-to-end system that trains a converged map over all 29.9 million MEDLINE articles in the 2026 baseline carrying at least five MeSH descriptors on one 24 GB GPU, remains the only implementation of those tested that runs at 262,144 neurons on that card (5.5), and on a 141 GB H200 reaches 1,048,576 neurons - the largest SOM I am aware of, exceeding the WEBSOM patent map on both corpus and map size (5.6).\end{itemize}
\begin{itemize}\item \sloppy A scaling study showing that held-out quantisation error follows a smooth power law with no elbow across three decades of map size (5.6), and a matched-work efficiency comparison placing SparseBin.SOM $\sim$82$\times$ ahead of MedSOM and, at 128$\times$128, 621$\times$ ahead of the best multicore-CPU library, a ratio that climbs steeply with map size - end-to-end figures that bundle layout, update rule and precision, the isolated layout effect being the 4.5-8.5$\times$ of 5.1 (5.5).\end{itemize}
The remainder of the paper reviews background and related work (2), develops the method (3), describes the experimental setup (4), presents results (5), and discusses implications, limitations, and reproducibility (6-9).

\section{Background and related work}
\subsection{Self-organizing maps and quality measures}
I use the batch SOM \cite{kohonen2001,kohonen2013}. Each training epoch proceeds in two phases: an assignment phase that maps every input to its BMU, and an update phase that replaces each neuron's prototype with the weighted mean of the inputs assigned to units in its neighbourhood, under a neighbourhood kernel that shrinks over training. I report three standard quality measures \cite{vesanto2000}. Quantisation error (QE) is the mean distance between inputs and their BMU prototypes and measures how faithfully the codebook represents the data. Topographic error (TE) is the fraction of inputs whose first and second BMUs are non-adjacent on the lattice and measures how well the map preserves topology. The dead-unit fraction is the proportion of neurons that win no inputs. I compute QE under cosine distance for cross-implementation comparison and report the implementation's native Euclidean QE where relevant.

\subsection{GPU and parallel SOM implementations}
The package somoclu \cite{wittek2017}, the strongest openly available multicore-CPU baseline for this task, provides dense CPU, dense GPU, and sparse CPU kernels. Its GPU kernel is dense-only: at a vocabulary of $\sim$30,000 features a dense M$\times$V codebook and dense input representation are infeasible (the codebook alone would run to petabytes across a map-size sweep), so the only somoclu kernel viable for MEDLINE is the multicore sparse-CPU kernel. Prior GPU SOMs, including our own earlier CUDA implementation (MedSOM;~\cite{amos2024a,amos2024b}), accelerate the BMU search but use per-node update loops and node-major codebooks; I use MedSOM as a same-task GPU baseline in 5.5. A further distinction concerns how the best-matching unit is computed: most implementations - the classic Kohonen line, somoclu, and the cuSPARSE baseline, as well as the main model developed here, SparseBin.SOM - evaluate the exact Euclidean distance, carrying each node's $\|$w$_i$$\|$$^{2}$ and leaving the codebook un-normalised, so their prototypes remain interpretable cell means and their quantisation error is directly comparable. MedSOM is the exception, L2-normalising its codebook once per epoch so the search collapses to a maximum dot product - faster, but it shifts the effective metric toward cosine and renders its quantisation error non-comparable; this is why, in 5, the exact-argmin implementations agree on quality and only MedSOM's QE is excluded from the comparison.

\subsection{Sparse kernels, memory layout, and the roofline}
The BMU search is a product between the sparse input matrix and the dense codebook, closely related to SpMM. Library SpMM (cuSPARSE) exposes a row- versus column-major output ordering (ORDER\_ROW vs ORDER\_COL) that corresponds directly to the feature-major versus node-major codebook distinction, letting me isolate the layout effect within a single library (5.2). Because the kernel is memory-bound, the relevant analytical frame is the roofline model \cite{williams2009}: performance is set by achieved bandwidth and arithmetic intensity rather than peak FLOPs, which motivates a layout that maximises coalesced reads and data reuse.

\subsection{How large can a SOM be, and how is its size chosen?}
The largest SOM in the peer-reviewed literature is the WEBSOM culmination of Kohonen et al. (2000), which mapped 6,840,568 patent abstracts - represented as 500-dimensional random projections of weighted word histograms - onto a 1,002,240-node map. Two points are relevant here. First, the map size was not derived from any optimality metric: it was set by what was computationally feasible and by the browsing use-case, and the large grid was reached by an engineering shortcut - training a much smaller grid first, then inserting interstitial nodes and interpolating their weights. Quantisation error appears in that work as a quality justification, not as a size selector. Second, the only project operating at this scale therefore sized its map by budget and task, not by a breakpoint in the data.

The literature that does claim a useful size criterion operates orders of magnitude smaller. The most principled recent proposal, SMLSOM \cite{motegi2023}, selects the number of units automatically by deleting redundant ones under a minimum-description-length criterion, but is demonstrated on small real datasets. The growing/hierarchical family - the Growing Hierarchical SOM \cite{rauber2002} and descendants such as RT-GSOM \cite{pramanik2021} - is evaluated on UCI-repository and small text collections (roughly 10$^3$-10$^5$ samples). Notably, these methods sidestep the flat-map size question rather than answer it: a GHSOM does not select a single optimal flat size but grows a hierarchy in which each region expands until its local quantisation error falls below a fraction of its parent's, governed by a local threshold rather than a global QE/TE elbow. In short, optimality-by-metric has been demonstrated only at benchmark scale, while the one $\sim$10$^7$-sample SOM was sized by resources - context that frames the scaling result in 5.6.

More directly, the present work scales our own prior MEDLINE SOMs. A 350$\times$350 (122,500-node) batch SOM of the MEDLINE corpus was used to build a cartographic atlas of medical knowledge \cite{amos2024a}, its map size selected from a topographic-error plateau. It was subsequently validated by projecting an expert-derived textbook knowledge structure - the reference lists of ten editions of a standard psychiatric textbook - onto the trained map \cite{amos2024b}. That 350$\times$350 map is the MedSOM baseline of 5.5; the feature-major method carries the same modelling programme from the $\sim$10$^5$-neuron regime to $\sim$10$^6$ neurons, and the companion validity paper \cite{amosprep} extends the external-validation argument of Amos et al.~\cite{amos2024b} to the MeSH tree and citation-derived taxonomies.

\section{Method}
\subsection{Problem statement and notation}
\begin{table}[htbp]\centering
\caption{Notation used throughout.}
\label{tab:notation}
\small
\begin{tabular}{ll}
\toprule
Symbol & Meaning \\
\midrule
$N$ & number of input articles \\
$V$ & number of MeSH features (vocabulary size) \\
nnz & nonzero features per input (MeSH per article) (mean = 11.1) \\
$M$ & number of neurons (map units), $M=E^{2}$ \\
$E$ & edge of the square lattice \\
$i$, $v$ & neuron index, feature index \\
$W$ & codebook comprising $M\cdot V$ weights \\
$x$, $w_i$ & input vector, prototype of neuron $i$ \\
TA & articles per BMU tile (16 in production) \\
$\sigma$ & neighbourhood radius \\
QE, TE & quantisation error, topographic error \\
\bottomrule
\end{tabular}

\vspace{0.6ex}
{\footnotesize\raggedright The BMU product contracts over $V$, not over the neuron count: the score block is $X_{N\times V}\,W^{\top}_{V\times M}$. Beware the BLAS collision - in GEMM's own $M{\times}K{\times}N$ naming, GEMM's $M$ is this paper's $N$, GEMM's $K$ is $V$ in this paper, and GEMM's $N$ is $M$ here.\par}

\end{table}

Let the corpus contain N inputs over V binary features, with each input x having a small number nnz of nonzero features (mean 11.1 in the present corpus). The map has M neurons arranged on a planar square lattice of edge E, so M = E$^{2}$. Each neuron $i$ carries a prototype $w_i \in \mathbb{R}^{V}$. The BMU of x is

\begin{equation}
\mathrm{BMU}(x)\;=\;\arg\min_{i}\;\lVert x-w_i\rVert^{2}
\;=\;\arg\min_{i}\;\bigl(\lVert w_i\rVert^{2}-2\langle x,w_i\rangle\bigr),
\label{eq:bmu}
\end{equation}

since $\|$x$\|$$^{2}$ is constant across neurons. For binary x the inner product $\langle$x, w$_i$$\rangle$ is a gather-sum over the input's nonzero features, so the BMU search reduces to, for each input, gathering the corresponding rows of the codebook and accumulating per-neuron partial sums and norms. To guarantee bit-reproducible results I sort each input's features in ascending order (fixing the floating-point accumulation order), compute $\|$w$_i$$\|$$^{2}$ in the same order, and break ties by lowest neuron index. Under this exact-match design every kernel and layout returns identical BMUs at a fixed tile size, which is the condition under which every comparison reported here was made. Changing the tile size changes the order in which each sample's score is accumulated, and roughly one assignment in 10$^{5}$ then differs at 256$\times$256 by a floating-point tie between grid-adjacent neurons; accumulating in double precision removes the difference entirely (8).

\subsection{The feature-major codebook}
I store the codebook feature-major: W[v$\cdot$M+i] places the M weights for feature v in a contiguous block. The consequence for the BMU search is that, when a warp processes the contribution of a single nonzero feature v, the M weights it must read are contiguous, so the loads coalesce into a small number of cache-line-aligned transactions. The node-major alternative W[i$\cdot$V+v] places a neuron's V weights contiguously; a warp computing different neurons for the same feature then issues a strided gather across the full vocabulary stride, and because the codebook far exceeds the L2 cache, each gather is an uncoalesced HBM round-trip. Beyond coalescing, the feature-major layout enables reuse: the same feature column, once resident, serves every input in a processing tile that contains that feature (3.3). The codebook holds M$\cdot$V weights; I store them in IEEE half precision (FP16, ``\_\_half'') and convert to FP32 only for the dot-product accumulation, so the resident codebook is 2$\cdot$M$\cdot$V bytes. Because the BMU is an exact argmin, halving the storage precision halves the codebook footprint with no effect on the assignment. This FP16 codebook is what lets the largest maps fit: the 262,144-neuron codebook occupies $\approx$16 GB in FP16, within the 4090's 24 GB where both cuSPARSE layouts run out of memory (5.5), and the 1,048,576-neuron codebook $\approx$65 GB, within the H200's 141 GB (5.6). It is worth noting that the benefit is layout-dependent - the feature-major search gains 1.65$\times$ from FP16, the node-major search only 1.07$\times$ (5.1).

\begin{figure}[htbp]\centering
\includegraphics[width=\linewidth]{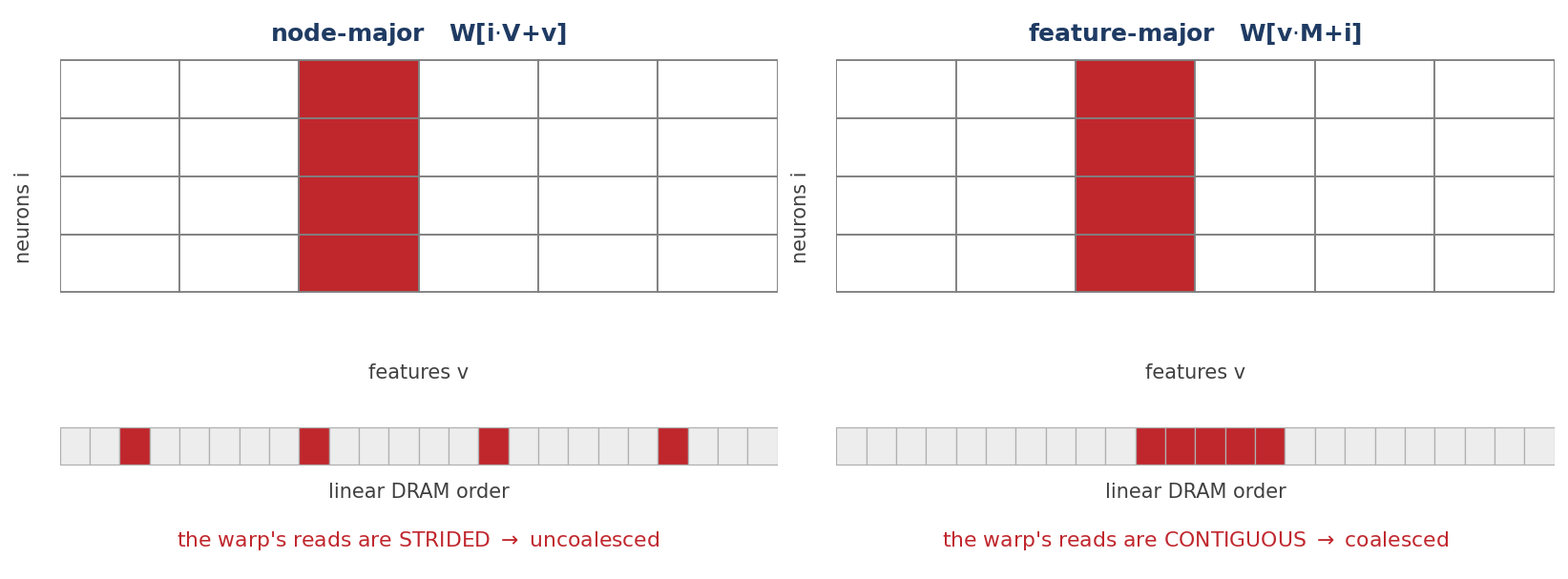}
\caption{Codebook layout and the BMU access pattern. Highlighted are the M weights one warp reads for a single feature v. Node-major storage (W[i$\cdot$V+v]) places a neuron's features contiguously, so reading one feature across neurons is strided in DRAM and the accesses cannot coalesce; feature-major storage (W[v$\cdot$M+i]) places a feature's neurons contiguously, so the same reads are a single coalesced transaction - and the loaded column can be reused across a tile.}
\label{fig:1}
\end{figure}

\subsection{The SpMM-tile BMU kernel}
The BMU kernel processes inputs in tiles of TA samples, TA being the tile extent along the sample axis of the product (the production kernel uses TA = 16; the microbenchmark of 5.1 used an earlier TA = 8). For each tile, the block first forms the union of the distinct features the tile's inputs touch - at most min(TA$\cdot$nnz, V) features, roughly 180 at TA = 16 in the present setting. It then sweeps these feature columns once: each loaded weight is applied to all TA inputs that contain that feature, accumulating per-neuron partial dot-products and norms, and the argmin (BMU selection) is fused into the same sweep rather than run as a separate pass. The read complexity is therefore O((N/TA)$\cdot$nnz$\cdot$M + N$\cdot$M), tracking the sparse lower bound N$\cdot$nnz$\cdot$M rather than the dense N$\cdot$V$\cdot$M of a naive sweep. The tile size trades occupancy against per-tile work: smaller tiles create more blocks and better parallelism, larger tiles increase reuse but enlarge the per-tile feature union.

\begin{figure}[htbp]\centering
\includegraphics[width=\linewidth]{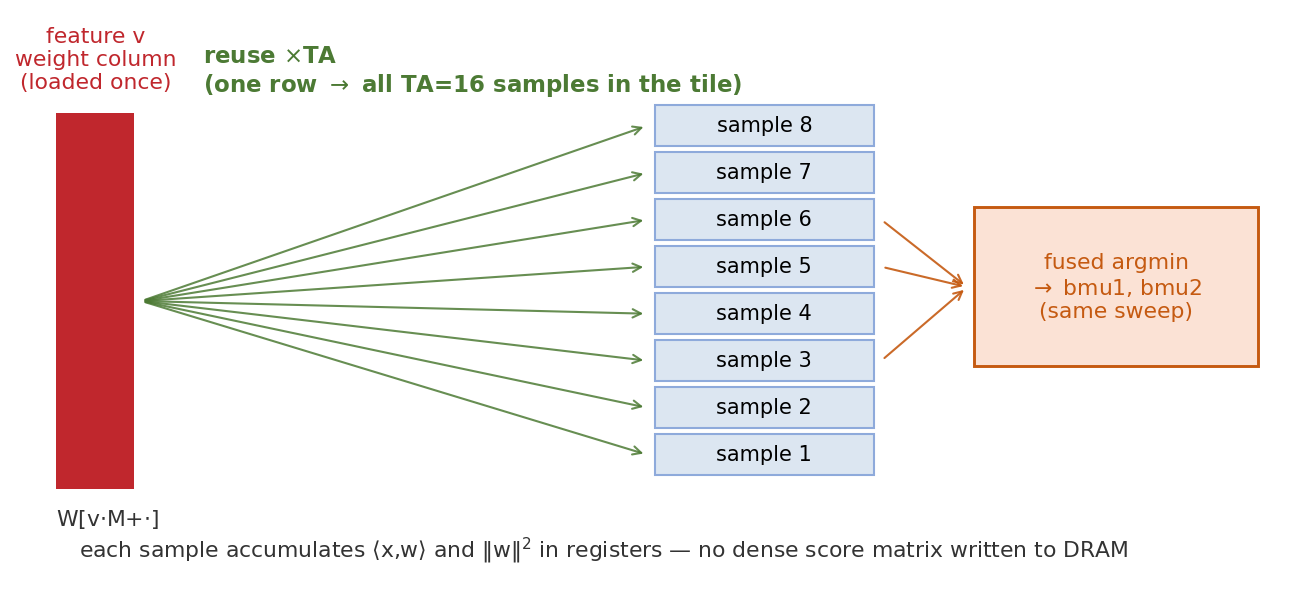}
\caption{The SpMM-tile BMU kernel. Each feature column, loaded once, is reused across all TA samples of a tile; every sample accumulates its dot-product and norm in registers, so no dense N$\times$M score matrix is written to DRAM (the round trip that dominates the cuSPARSE path, 5.7). The two best units per sample are found in the same fused sweep.}
\label{fig:2}
\end{figure}

\subsection{Factorised box-blur update, PCA initialisation, and the stopping rule}
The neighbourhood update is best seen as a convolution. Once the batch has been scattered onto the lattice - accumulating, per cell, the sum of the assigned samples and their count - replacing each neuron's prototype with its neighbourhood-weighted mean amounts to blurring those two accumulator fields with the neighbourhood kernel and dividing one by the other. I perform the blur as a factorised (separable) box filter: a one-dimensional pass along the rows followed by one along the columns. Each pass is maintained as a running window sum, so every cell costs a constant two operations - add the element entering the window, subtract the one leaving it - regardless of the window's width \cite{crow1984}. A box pass is therefore O(M) and, crucially, independent of the neighbourhood radius $\sigma$: widening the neighbourhood only moves the running sum's endpoints, it never adds work per cell. Three successive box passes approximate a Gaussian neighbourhood by the central-limit theorem \cite{wells1986}, so a fixed three passes in each direction reproduce a Gaussian-shaped kernel at fixed cost. The window half-width is $\sigma$ rounded to whole cells and never smaller than one, and three passes of radius $r$ give a kernel whose effective standard deviation is about $r+\tfrac12$ cells: $\sigma$ names the schedule variable, and the realised width is that much larger. This is what keeps the update flat in $\sigma$, and hence in map size (5.4). A direct Gaussian update costs about O(M$\cdot$$\sigma$$^{2}$) because it re-sums the whole neighbourhood area at every neuron, and even a separable Gaussian costs O(M$\cdot$$\sigma$), whereas the running-sum box blur is O(M) for any $\sigma$. The saving is largest exactly where it matters, since $\sigma$ starts near half the map edge and is largest on the biggest maps.

Four details matter for reimplementation. The box sums are unnormalised: the $(2r+1)$ constant cancels because the numerator and denominator fields receive the identical kernel, so an implementation that normalises one and not the other is silently wrong. The lattice boundary is clamped rather than toroidal, the window truncating at the edge identically in both fields, which leaves the ratio a proper weighted mean over the cells that remain. A neuron whose blurred denominator is zero is left unchanged rather than filled, which happens only when no sample landed anywhere within its neighbourhood support. And the accumulation and the blur are carried in FP32, the conversion to the FP16 codebook occurring only at the final write. The update itself is a pure replacement with no learning rate: every sample of the batch is weighted equally within the neighbourhood kernel each epoch.

The radius follows an exponential endpoint schedule from $\sigma$$_0$ = 0.5$\cdot$E down to a small $\sigma_{\min}$ (rate 0.3). 

I initialise the codebook by projecting onto the leading principal components of the input (PCA initialisation). Under this schedule the progress variable is the epoch index itself, so the epoch count is set by the schedule and the plateau rather than by the initialisation: within SparseBin.SOM, PCA and random initialisation reach the plateau within two epochs of each other at every map size, with no systematic direction, and at the default initial radius $\sigma$$_0$ = 0.5$\cdot$E to statistically indistinguishable held-out quality. That equivalence is a property of an epoch-driven schedule and does not transfer to implementations whose radius is driven by measured progress (5.2). 

PCA's value is therefore twofold and distinct from a raw epoch saving: it yields a deterministic, bit-reproducible result, and it licenses a smaller initial neighbourhood. Halving $\sigma$$_0$ to 0.25$\cdot$E under PCA improves held-out quantisation error by roughly 1\% (up to about 2\% on the largest maps) while trimming a few epochs, whereas the same reduction degrades quality under random initialisation, which depends on the wide-neighbourhood ordering phase that a pre-ordered PCA map renders largely redundant. 

Batch SOM training has no intrinsic length - too few epochs leave the map under-organised, too many merely waste compute - and the natural number depends on map size, initialisation, and the annealing schedule, so any fixed epoch budget is arbitrary. Training instead passes through two regimes: a broad ordering phase (large $\sigma$) that fixes the global topological arrangement of the lattice, and a fine-tuning phase (small $\sigma$) in which that arrangement is settled and only local quantisation still improves. 

In the fine-tuning phase a plateau in map distortion is a reliable indicator of convergence \cite{kaski1996}, so I stop when it flattens rather than after a preset number of epochs. The monitored quantity is the Kaski-Lagus distortion: for each monitored sample, the distance from the input to its best-matching prototype plus the length of a lattice path from the best to the second-best unit, each edge of that path measured as the input-space distance between adjacent prototypes and the path capped at eight edges; the reported value is the mean over the monitored set. 

Training stops once the map has entered the refinement regime ($\sigma \leq \max(1,\,2\sigma_{\min}) = 1$, at every map size) and the relative change in that quantity stays below 0.1\% (a relative change < 0.001) for three consecutive epochs, which under the configuration used throughout occurs after 20 to 29 epochs depending on map size. A separate check that topographic error is at most 0.5 decides whether a run is labelled converged; it does not affect when training halts. That is what makes the epoch count adapt to each map rather than being fixed in advance (contrast MedSOM's fixed 20 epochs, 5.5).

This brings me to the observation that underpins the evaluation. Because the BMU is an exact argmin, it is identical for any correct implementation and any codebook layout; the assignment phase cannot, even in principle, change clustering quality. Every difference in QE, TE, or dead-unit fraction between two SOM implementations must therefore come from the update axis - the neighbourhood kernel and $\sigma$ schedule - not from the BMU. I exploit this in 5.3 to show that, once a baseline is given the box-blur update, its quality converges to that of SparseBin.SOM, confirming that the layout/BMU speed-up costs nothing in quality.

\begin{table}[p]\centering\footnotesize
\begin{tabular}{@{}rp{0.78\linewidth}l@{}}
\toprule
\multicolumn{3}{@{}l}{\textbf{Algorithm 1   MedSOM batch SOM - node-major, as published \cite{amos2024b}}} \\
\midrule
Input: & X - N sparse-binary vectors over V features; M = E$^{2}$ neurons (planar square lattice) &  \\
\rowcolor{noteA}Data: & W - codebook of M dense node vectors, node-major W[i$\cdot$V+v], FP32 & a \\
1 & randomise W &  \\
\rowcolor{noteD}2 & for epoch e = 1 \ldots{} 20 do & d \\
3 & \hspace{1.3em}$R = \max(1,\ \lfloor (E/2)\,/\,1.7^{\,e-1} \rfloor)$   // neighbourhood radius; $E/2$ at $e=1$. Gaussian width is $R/2$ &  \\
\rowcolor{noteB}4 & \hspace{1.3em}L2-normalise every node vector of W & b \\
\rowcolor{noteA}5 & \hspace{1.3em}for each sample, each neuron i: dot  $\langle$x,w$_i$$\rangle$   // node-major strided gather & a \\
\rowcolor{noteB}6 & \hspace{1.3em}\hspace{1.3em}rank units by $\|$x$\|$$_0$ $-$ 2$\cdot$dot$_i$   // $\|$w$_i$$\|$$^{2}$ omitted: step 4 makes it 1 for every unit & b \\
7 & \hspace{1.3em}\hspace{1.3em}bmu1, bmu2  two smallest &  \\
8 & \hspace{1.3em}\hspace{1.3em}accumulate per-node numerators / denominators &  \\
\rowcolor{noteC}9 & \hspace{1.3em}W  weighted mean, Gaussian $l(e)=\exp(-\lVert r_i-r_c\rVert_1^{2}/2(R/2)^{2})$   // no radius cutoff: every node sweeps every article, so cost is independent of $R$ & c \\
10 & \hspace{1.3em}TE  fraction with bmu1, bmu2 non-adjacent &  \\
11 & end for &  \\
\rowcolor{noteD}12 & return the codebook with the lowest TE over the 20 epochs & d \\
\bottomrule
\end{tabular}

\vspace{1.6ex}

\begin{tabular}{@{}rp{0.78\linewidth}l@{}}
\toprule
\multicolumn{3}{@{}l}{\textbf{Algorithm 2   SparseBin.SOM: feature-major sparse-binary batch SOM}} \\
\midrule
Input: & X - N sparse-binary vectors over V features; M = E$^{2}$ neurons (planar square lattice) &  \\
\rowcolor{noteA}Data: & W - codebook of M$\times$V weights, feature-major W[v$\cdot$M+i], FP16 & a \\
1 & initialise W by projection onto the leading principal components (PCA) &  \\
\rowcolor{noteD}2 & for epoch e = 1, 2, \ldots{} until the Kaski-Lagus plateau do & d \\
3 & \hspace{1.3em}$\sigma$  exponential endpoint schedule $\sigma$$_0$ = 0.5$\cdot$E $\rightarrow$ $\sigma_{\min}$ &  \\
\rowcolor{noteA}4 & \hspace{1.3em}for each tile of TA samples do          // SpMM-tile: coalesced + reuse & a \\
5 & \hspace{1.3em}\hspace{1.3em}load the distinct feature columns the tile touches &  \\
\rowcolor{noteB}6 & \hspace{1.3em}\hspace{1.3em}for each neuron i: accumulate $\langle$x,w$_i$$\rangle$ and $\|$w$_i$$\|$$^{2}$   // exact, keeps $\|$w$_i$$\|$$^{2}$ & b \\
7 & \hspace{1.3em}\hspace{1.3em}bmu1, bmu2  two smallest ($\|$w$_i$$\|$$^{2}$ $-$ 2$\langle$x,w$_i$$\rangle$)   // fused argmin &  \\
8 & \hspace{1.3em}\hspace{1.3em}accumulate num, den (feature-major) &  \\
\rowcolor{noteC}9 & \hspace{1.3em}blur num and den, three box passes per axis of half-width $\lfloor\sigma\rceil$;  W  num / den   // $\sigma$-independent & c \\
10 & \hspace{1.3em}QE  mean(1 $-$ cos(x,w$_{\mathrm{bmu1}}$));  TE  fraction non-adjacent &  \\
\rowcolor{noteD}11 & \hspace{1.3em}if $\sigma \leq 1$ and $\Delta$KL $<\varepsilon$ for three consecutive epochs then break   // KL = Kaski-Lagus distortion & d \\
12 & return the trained codebook W and the BMU of every sample &  \\
\bottomrule
\end{tabular}
\end{table}
\clearpage

It is instructive to set the two batch SOMs side by side. Algorithms 1 and 2 share the same skeleton - anneal a neighbourhood, assign every sample to its best-matching unit, move each neuron toward the mean of its neighbourhood, repeat - but differ in four places that, between them, account for essentially all of the speed, scaling, and reproducibility differences reported in 5. The steps that differ are shaded and keyed a-d to the notes beneath the algorithms, with the same tint used for the same note in both algorithms so that corresponding steps line up; the unshaded steps are common to both.

Algorithm 1. MedSOM, the node-major MEDLINE batch SOM of the earlier published work \cite{amos2024b}. Shaded steps, keyed a-d, are those that differ from Algorithm 2; tints match across the two algorithms. Step 5 is the strided, uncoalesced gather drawn in the left-hand panel of Figure 1.

Algorithm 2. The feature-major method introduced here. Shaded steps, keyed a-d, mark the four high-impact differences from Algorithm 1; tints match Algorithm 1. Step 4 is the coalesced feature-column read drawn in the right-hand panel of Figure 1, and steps 4-7 are the tile sweep of Figure 2, in which one loaded column serves every sample of the tile and the two best units are taken without materialising a score matrix. See the notes beneath.

 a  Memory layout and BMU access - the decisive performance change. MedSOM stores the codebook node-major and gathers it with a strided, uncoalesced pattern that thrashes the L2 cache; the feature-major layout makes each feature column a coalesced read and reuses it across the TA samples of a tile. This single change is worth 4.5-8.5$\times$ on the BMU search when nothing else is varied (5.1), and is the reason the BMU dominates training time at every scale (5.4).

 b  Distance calculation. MedSOM L2-normalises the whole codebook every epoch so it can rank units by $\|$x$\|$$_0$ $-$ 2$\langle$x,w$\rangle$, dropping $\|$w$\|$$^{2}$; this normalisation is destructive (the learned weights are rescaled each epoch) and renders its QE non-comparable across implementations. Algorithm 2 carries an exact $\|$w$_i$$\|$$^{2}$ in the fused BMU sweep, so the codebook is preserved and QE is directly interpretable.

 c  Neighbourhood update. MedSOM applies its Gaussian with no radius cutoff, so every node sweeps every article on every epoch and its per-epoch cost is independent of the radius and maximal at all radii. The cuSPARSE baseline's Gaussian does scale with the radius, growing 133-fold across a sixty-fourfold increase in neurons, whereas the factorised three-pass box blur is $\sigma$-independent and grows eightfold. This is why the update advantage grows with map size - from parity at 32$\times$32 to 14.5$\times$ at 256$\times$256 (5.4).

 d  Stopping rule. MedSOM runs a fixed 20 epochs and keeps the lowest-TE codebook post hoc; Algorithm 2 stops adaptively on a Kaski-Lagus plateau rather than after a fixed epoch budget (3.4).

Fidelity of Algorithm 1 to MedSOM. Algorithm 1 describes both the published MedSOM \cite{amos2024b} and the implementation benchmarked here: the two share the same training-path source, the port having touched only input handling and build plumbing. Three details are compressed for readability. First, the radius schedule carries a floor at 1 and is truncated to an integer, so at $E=350$ the sequence over 20 epochs is 175, 102, 60, 35, 20, 12, 7, 4, 2 and then 1 for the remaining eleven epochs. Second, the published schedule is printed as $175/(1.7)^{\text{epoch}}$ \cite{amos2024b}, but the epoch counter is zero-indexed in the source, so the first epoch uses the full $E/2$ rather than $E/2$ divided by 1.7; Algorithm 1 follows the code. 

Third, step 12's selection of the lowest-topographic-error codebook is performed post hoc over saved checkpoints rather than by the program itself, and the error that gates a checkpoint is measured before that epoch's update while the file is written after it. Separately, MedSOM runs in FP32 throughout, where Algorithm 2 stores its codebook in FP16 and accumulates in FP32; because the benefit of FP16 is layout-dependent, the consequences for the timing comparison are bounded rather than removed, and are quantified in 5.5.

\section{Experimental setup}
\subsection{Hardware}
All comparisons between implementations were measured on a single NVIDIA RTX 4090 (AD102; 24 GB VRAM, 72 MB L2, $\sim$1 TB/s peak bandwidth, 82.6 TFLOP/s FP32), driver 610.43.02, CUDA 12.8. The largest map in the size sweep (1024$\times$1024, 5.6) was trained on an H200 SXM (141 GB VRAM) rented on demand, under the earlier adaptive $\sigma$ schedule; it is labelled as such wherever it appears. The somoclu CPU baseline (version 1.7.6, build 1.7.6-11-g63895f4, sparse CPU kernel) runs on the host over 28 threads. Provenance - submodule commits, resolved configuration hash, GPU, driver, and CUDA version - is recorded per artifact by the reproduction pipeline.

Use of AI tools in the research process. Generative AI assistance (Anthropic Claude) was used throughout the research process: implementing and running the training and profiling experiments, building the reproduction pipeline, writing the analysis and figure-generation scripts, and performing the statistical analyses reported in 5. All figures are derived directly from the released data by reproducible scripts, which are part of the code release (8); no figure contains AI-generated content that is not computed from the underlying measurements. Every experimental design decision, every analysis choice, and every claim made in this paper was directed and approved by the author, who takes responsibility for them.

\subsection{Corpus and data format}
Articles were drawn from the PubMed 2026 baseline, a fixed snapshot, and retained only if they carried at least five MeSH descriptors, which yields 29,903,261 articles over 30,766 descriptors. The resulting corpus is archived at \url{https://doi.org/10.5281/zenodo.20770707} (8). Records outside the MeSH-indexed MEDLINE subset - PubMed Central deposits, in-process and ahead-of-print citations, books - carry no descriptors and are therefore never candidates; the floor of five is what additionally excludes thinly indexed MEDLINE records. The mean of 11.1 descriptors per article follows from that floor. The corpus is stored in .sbcsr (sparse-binary compressed-sparse-row), a bespoke on-disk container introduced for this work rather than a community standard; the layout inside it is ordinary CSR, with the value array omitted because every stored element is 1. Each article is a row holding the sorted integer indices of the MeSH descriptors present in it, in the standard CSR layout - a row-pointer array giving where each article's index list begins, plus one concatenated array of those descriptor indices. The entry values are left implicit, because in a binary incidence matrix every stored element is a 1. Recording presence alone (about eleven indices per article) rather than a dense 30,766-element vector, the entire 29.9-million-article corpus occupies roughly 750 MB on disk and streams directly into the GPU kernels without decompression.

\begin{table}[htbp]\centering
\caption{Corpus statistics.}
\label{tab:1}
\small
\begin{tabular}{@{}>{\raggedright\arraybackslash}p{0.30\linewidth}>{\raggedright\arraybackslash}p{0.62\linewidth}@{}}
\toprule
Property & Value \\
\midrule
Source & PubMed 2026 baseline (1,334 files, complete) \\
Filter & Articles with $\geq$ 5 MeSH descriptors \\
Articles (N) & 29,903,261 \\
Vocabulary (V) & 30,766 MeSH descriptors \\
Nonzeros & 332,436,043 (mean 11.12 per article, sd 4.60; density $\approx$ 0.036\%) \\
Format & .sbcsr sparse-binary CSR; anonymised (no PMIDs); released under CC0 at doi:10.5281/zenodo.20770707 \\
\bottomrule
\end{tabular}

\vspace{0.6ex}
{\footnotesize\raggedright\par}

\end{table}

\subsection{Baselines and comparators}
Four implementations are compared, and are named in the tables as follows. \textbf{SparseBin.SOM} is the feature-major sparse-binary method introduced here (Algorithm 2), and \textbf{SparseFloat.SOM} is the same design over real-valued sparse inputs. \textbf{cuSPARSE.SOM} is the baseline described below, reported in whichever codebook ordering is relevant and labelled feature-major or node-major accordingly. \textbf{MedSOM} is the prior CUDA implementation of Algorithm 1, and \textbf{somoclu} is the multicore CPU library. The repositories that hold these implementations carry their development names rather than the labels used here, and the release README gives the correspondence.

\begin{itemize}\item \sloppy cuSPARSE.SOM - a same-GPU cuSPARSE baseline (cusparseSpMM; \cite{nvidia_cusparse}), run in both feature-major (ORDER\_ROW) and node-major (ORDER\_COL) orderings.\end{itemize}
\begin{itemize}\item MedSOM - our prior CUDA SOM, ported to Linux with native .sbcsr loading; serial per-node update and an approximate Euclidean metric (it L2-normalises the codebook each epoch, then ranks by nnz $-$ 2$\cdot$dot).\end{itemize}
\begin{itemize}\item somoclu - the multicore sparse-CPU kernel, the only somoclu kernel viable at V $\approx$ 30k.\end{itemize}
Fairness is controlled by using the identical corpus and map sizes, matching the per-size epoch budget and $\sigma$ endpoints where the comparison is on training cost, and scoring all codebooks with a single shared cosine evaluator. Documented asymmetries - kernel shape (box-blur vs Gaussian), GPU vs CPU, initialisation (PCA here, uniform random in the cuSPARSE baselines, which have no PCA option), and the driver of the $\sigma$ schedule (the epoch index here, measured progress in the baselines) - are disclosed rather than hidden, and are noted where they bear on a specific comparison.

\subsection{Quality metrics}
\sloppy Cross-implementation quality is scored by: QE = mean 1 $-$ cos(x, BMU$_1$); TE = fraction of inputs whose first and second BMUs are non-adjacent (Chebyshev distance > 1); and the dead-unit fraction. Every implementation is scored the same way, by a single external evaluator reading its trained codebook, rather than by trusting the error each implementation prints for itself. This matters because those self-reported figures are computed under whatever internal convention the implementation happens to use: MedSOM, for instance, L2-normalises its codebook every epoch, so its internal quantisation error is measured against rescaled weights on a metric that has drifted toward cosine, and is numerically incomparable with an error computed from unnormalised prototypes. Self-reported values are therefore used only to track progress within a single implementation, never to compare one implementation against another.

\section{Results}
\subsection{Codebook layout: feature- versus node-major}
The cleanest test of the layout argument holds everything else fixed. Because the cuSPARSE baseline exists in both layouts - the same implementation, the same precision, the same update and stopping rule, differing only in whether the codebook is stored node-major or feature-major - the ratio between them isolates the effect of memory layout alone (Table~\ref{tab:2}). Feature-major storage accelerates the best-matching-unit search by 8.5$\times$ at 32$\times$32, and the advantage remains between 4.5$\times$ and 8.1$\times$ across the rest of the ladder. The mechanism is the one set out in 3.2: reading one feature's weights across neurons is a coalesced transaction under feature-major storage and a strided gather under node-major, and at a codebook far larger than the L2 cache every uncoalesced gather becomes an HBM round trip. That mechanism holds from 128$\times$128 upward. At 32$\times$32 the FP16 codebook is 63 MB against this card's 72 MB of L2 and is cache-resident, so the layout penalty at that rung is an access-pattern effect within L2 rather than a DRAM effect.

\begin{table}[htbp]\centering
\caption{Effect of codebook layout in isolation.}
\label{tab:2}
\small
\resizebox{\textwidth}{!}{%
\begin{tabular}{lrrr}
\toprule
Edge & Node-major BMU/epoch (s) & Feature-major BMU/epoch (s) & Layout speed-up \\
\midrule
32 & 4.04 & 0.478 & 8.5$\times$ \\
64 & 16.31 & 2.02 & 8.1$\times$ \\
128 & 65.65 & 8.72 & 7.5$\times$ \\
256 & 289.66 & 63.84 & 4.5$\times$ \\
\bottomrule
\end{tabular}}

\vspace{0.6ex}
{\footnotesize\raggedright cuSPARSE.SOM in its node-major and feature-major variants, full corpus, matched update (box+KL), FP16, mean of five seeds. The two columns differ only in codebook layout, so the ratio is attributable to layout alone.\par}

\end{table}

A second, independent observation confirms that layout - not precision - is what limits the node-major path. Halving the codebook to FP16 accelerates the feature-major kernel by 1.65$\times$, close to the 2$\times$ its halved read volume would predict, but accelerates the node-major kernel by only 1.07$\times$. The node-major search is not bandwidth-bound at all: it is limited by the latency of dependent, uncoalesced gathers, so giving it fewer bytes to move barely helps. For that layout the access pattern is the whole problem, which is why I treat the layout rather than the storage format as the primary design decision.

\subsection{Equal-quality comparison with cuSPARSE: a crossover, and a capacity wall}
I next compare SparseBin.SOM against the stronger of the two cuSPARSE.SOM baselines at matched quality, letting each run to its own Kaski-Lagus plateau rather than a fixed budget (Table~\ref{tab:3}). Held-out quantisation error agrees to within 0.5\% at every map size, and to within 0.07\% at the two largest - so whatever else differs, the two implementations are converging to maps of the same representational quality. On topographic error and dead-unit fraction SparseBin.SOM is slightly ahead (at 256$\times$256, TE 0.033 versus 0.037 and 10.7\% versus 17.4\% dead), so the comparison is equal-or-better rather than merely equal.

Against that fixed quality, the timing is not a single multiple but a crossover (Table~\ref{tab:cross}). Wall-clock time throughout this paper means elapsed real time, as distinct from summed kernel time or CPU time; where the scope is a whole training run, as here, it is the total from initialisation to the Kaski-Lagus plateau, summed over however many epochs that run needed rather than cost per epoch. The two differ because the runs converge in different numbers of epochs. At 32$\times$32 the cuSPARSE path is 2.8$\times$ faster; at 64$\times$64 the two are at parity; from 128$\times$128 onward SparseBin.SOM pulls ahead, by 1.5$\times$ at 128$\times$128 and 2.6$\times$ at 256$\times$256. Two effects compound in SparseBin.SOM's favour as the map grows: the per-epoch BMU search becomes faster (from 0.23$\times$ of the baseline's speed at 32$\times$32 to 1.53$\times$ at 256$\times$256), and the present runs converge in fewer epochs (25 against 40 at 256$\times$256). 

The epoch difference is not a property of the layout, and is worth attributing plainly. The two arms differ in three disclosed respects that bear on epoch count: SparseBin.SOM is PCA-initialised while the baseline draws its codebook uniformly at random; SparseBin.SOM anneals on an epoch-driven deterministic schedule while the baseline's radius is driven by measured progress, so it holds the radius wide for as long as the map keeps improving; and, more marginally, they evaluate the stopping metric over monitored sets of different size (the first 100,000 samples here, 200,000 in the baseline). Both arms test the same fully weighted Kaski-Lagus distortion, and the one remaining gate difference - the path-term guard - decides only whether a run is labelled converged, not when it halts. 

A random start leaves more ordering work to do, and under a progress-driven radius that work converts directly into epochs, which is why the gap widens with map size (one epoch at 32$\times$32, roughly fifteen at 256$\times$256). The per-epoch BMU advantage is attributable to layout; the epoch advantage is attributable to initialisation and schedule, and the wall-clock figures in Table~\ref{tab:cross} combine the two. At 512$\times$512 the comparison ends - both cuSPARSE layouts exhaust the 24 GB card, while SparseBin.SOM trains on. The honest summary is therefore that the bespoke kernel is not universally faster; it is faster where large maps are wanted, and it is the only one of the two that reaches them.

\begin{table}[htbp]\centering
\caption{The equal-quality crossover.}
\label{tab:cross}
\small
\resizebox{\textwidth}{!}{%
\begin{tabular}{@{}lrrrrr@{}}
\toprule
Edge & \multicolumn{2}{c}{Wall to plateau (s)} & \multicolumn{2}{c}{Epochs to plateau} & Speed-up \\
\cmidrule(lr){2-3}\cmidrule(lr){4-5}
 & cuSPARSE.SOM & SparseBin.SOM & cuSPARSE.SOM & SparseBin.SOM & (SparseBin.SOM) \\
\midrule
32  & 18.8    & 52.8    & 23 & 22 & 0.36$\times$ \\
64  & 75.5    & 71.8    & 27 & 20 & 1.05$\times$ \\
128 & 366.7   & 249.3   & 34 & 25 & 1.47$\times$ \\
256 & 2,758.2 & 1,061.5 & 40 & 25 & 2.60$\times$ \\
512 & OOM     & 5,427.1 & -  & 29 & -   \\
\bottomrule
\end{tabular}}

\vspace{0.6ex}
{\footnotesize\raggedright Wall-clock time from initialisation to each run's own Kaski-Lagus plateau, so the
totals absorb both per-epoch cost and the number of epochs required. A speed-up below 1 means cuSPARSE is
faster. The crossover falls between 32$\times$32 and 64$\times$64, and the margin then widens with map size,
driven by both a faster per-epoch BMU search and earlier convergence. At 512$\times$512 both cuSPARSE layouts
exhaust the 24 GB card, so no comparison is possible. Held-out QE agrees to within 0.5\% at every size
(Table~\ref{tab:3}).\par}
\end{table}

\begin{table}[htbp]\centering
\caption{Equal-quality comparison, full corpus.}
\label{tab:3}
\footnotesize
\resizebox{\textwidth}{!}{%
\begin{tabular}{lrrrrrrr}
\toprule
Edge & Model & Wall (s) & BMU/ep (s) & Epochs & QE (held) & TE & Dead \% \\
\midrule
32$\times$32 & SparseBin.SOM & 52.8 & 2.296 & 22 & 0.5241 & 0.010 & 9.4 \\
32$\times$32 & cuSPARSE.SOM & 18.8 & 0.478 & 23 & 0.5264 & 0.015 & 9.8 \\
64$\times$64 & SparseBin.SOM & 71.8 & 3.432 & 20 & 0.4899 & 0.018 & 8.7 \\
64$\times$64 & cuSPARSE.SOM & 75.5 & 2.024 & 27 & 0.4885 & 0.021 & 10.8 \\
128$\times$128 & SparseBin.SOM & 249.3 & 9.653 & 25 & 0.4518 & 0.027 & 9.3 \\
128$\times$128 & cuSPARSE.SOM & 366.7 & 8.722 & 34 & 0.4515 & 0.029 & 12.9 \\
256$\times$256 & SparseBin.SOM & 1,061.5 & 41.79 & 25 & 0.4152 & 0.033 & 10.7 \\
256$\times$256 & cuSPARSE.SOM & 2,758.2 & 63.84 & 40 & 0.4154 & 0.037 & 17.4 \\
512$\times$512 & SparseBin.SOM & - & 184.8 & - & - & - & - \\
512$\times$512 & cuSPARSE.SOM & OOM & OOM & - & - & - & - \\
\bottomrule
\end{tabular}}

\vspace{0.6ex}
{\footnotesize\raggedright All runs use the feature-major cuSPARSE baseline, matched update (box+KL) and FP16 throughout, each run stopped on its own Kaski-Lagus plateau; mean of five seeds. Held-out QE agrees within 0.5\% at every size (0.07\% at 128$\times$128 and 0.05\% at 256$\times$256). The wall-clock time advantage crosses over between 64$\times$64 and 128$\times$128; both cuSPARSE layouts exhaust 24 GB at 512$\times$512, where the 512$\times$512 BMU figure for SparseBin.SOM comes from the fixed-epoch efficiency sweep (5.5). BMU, best-matching unit; QE, quantisation error; TE, topographic error; OOM, out of memory (the run could not be allocated on the 24 GB card).\par}

\end{table}

\subsection{Quality is set by the update, not the BMU}
That the two implementations converge to the same quality is not a coincidence, and the reason is worth isolating. Giving the cuSPARSE baseline each of the update rules in turn (Table~\ref{tab:4}) shows that quality tracks the neighbourhood kernel and the stopping rule, not the search. Under a Gaussian neighbourhood the baseline has a topographic error between 0.15 and 0.23; replacing it with three box passes alone drops TE to 0.013-0.033, and adding the Kaski-Lagus stop brings quantisation error to within 0.5\% of SparseBin.SOM at every size. Because the best-matching unit is an exact argmin, identical in every one of these runs, the quality differences can only come from the update - exactly as the exact-argmin argument predicts. The layout speed-up of 5.1 is therefore free of any quality cost.

\begin{table}[htbp]\centering
\caption{Effect of the update rule on quality.}
\label{tab:4}
\footnotesize
\begin{tabular}{lrrrrr}
\toprule
Edge & Impl & Update & TE & QE (held) & Dead \% \\
\midrule
32 & cuSPARSE.SOM (feature-major) & Gaussian & 0.152 & 0.4889 & 3.5 \\
32 & cuSPARSE.SOM (feature-major) & box & 0.013 & 0.5289 & 9.5 \\
32 & cuSPARSE.SOM (feature-major) & box+KL & 0.015 & 0.5264 & 9.8 \\
32 & SparseBin.SOM & box+KL & 0.010 & 0.5241 & 9.4 \\
128 & cuSPARSE.SOM (feature-major) & Gaussian & 0.214 & 0.4222 & 3.9 \\
128 & cuSPARSE.SOM (feature-major) & box & 0.028 & 0.4601 & 12.8 \\
128 & cuSPARSE.SOM (feature-major) & box+KL & 0.029 & 0.4515 & 12.9 \\
128 & SparseBin.SOM & box+KL & 0.027 & 0.4518 & 9.3 \\
256 & cuSPARSE.SOM (feature-major) & Gaussian & 0.229 & 0.3896 & 6.2 \\
256 & cuSPARSE.SOM (feature-major) & box+KL & 0.037 & 0.4154 & 17.4 \\
256 & SparseBin.SOM & box+KL & 0.033 & 0.4152 & 10.7 \\
\bottomrule
\end{tabular}

\vspace{0.6ex}
{\footnotesize\raggedright Quality is governed by the update, not the BMU. The Gaussian rows show the baseline's native update; the box rows substitute the three-pass box blur; box+KL adds the Kaski-Lagus stop. TE falls by an order of magnitude with the box blur alone, and box+KL brings the baseline to within 0.5\% of the held-out QE here at every size. Mean of five seeds.\par}

\end{table}

\subsection{The update phase, and where training time actually goes}
The factorised box blur was introduced (3.4) to make the update cost independent of the neighbourhood radius, and the measurements bear this out (Table~\ref{tab:5}). The box-blur update rises from 0.031 to 0.248 seconds per epoch as the map grows from 32$\times$32 to 256$\times$256 - an eightfold increase for a sixtyfourfold increase in neurons - while the cuSPARSE path's update, which re-sums a neighbourhood whose area grows with $\sigma$$^{2}$, rises from 0.027 to 3.60 seconds. The bespoke update starts on par at the smallest map and is 14.5$\times$ cheaper by 256$\times$256, so the feature-major design does not merely avoid harming the update phase, it improves it, and the margin widens with map size exactly as the O(M) versus O(M$\cdot$$\sigma$$^{2}$) argument predicts.

\begin{table}[htbp]\centering
\caption{Neighbourhood-update cost per epoch.}
\label{tab:5}
\small
\begin{tabular}{lrrrr}
\toprule
Edge & \shortstack{SparseBin.SOM\\update (s/ep)} & \shortstack{cuSPARSE.SOM\\update (s/ep)} & Ratio & \shortstack{SparseBin.SOM\\BMU (s/ep)} \\
\midrule
32 & 0.031 & 0.027 & 0.9$\times$ & 2.296 \\
64 & 0.037 & 0.102 & 2.8$\times$ & 3.432 \\
128 & 0.082 & 0.558 & 6.8$\times$ & 9.653 \\
256 & 0.248 & 3.599 & 14.5$\times$ & 41.793 \\
\bottomrule
\end{tabular}

\vspace{0.6ex}
{\footnotesize\raggedright Matched update rule (box+KL), full corpus, mean of five seeds. The radius-independent box blur grows 8$\times$ across a 64$\times$ increase in neurons; the cuSPARSE path's grows 133$\times$. The rightmost column gives the BMU cost for scale - the update is a small fraction of it throughout.\par}

\end{table}

Set against the BMU search, the update is almost negligible (Figure 3). The best-matching-unit search consumes 98.7\% of training wall-clock time at 32$\times$32, rising monotonically to 99.5\% at 512$\times$512, while the update never exceeds 1.3\%. Both figures are properties of the configuration benchmarked here; at the tuned configuration of 7 the same measurement gives 73.7\% to 94.0\%, because the search was accelerated and the update was not. Optimising the codebook layout is therefore not one improvement among several; it is the only part of the computation whose cost matters at scale.

\begin{figure}[htbp]\centering
\includegraphics[width=\linewidth]{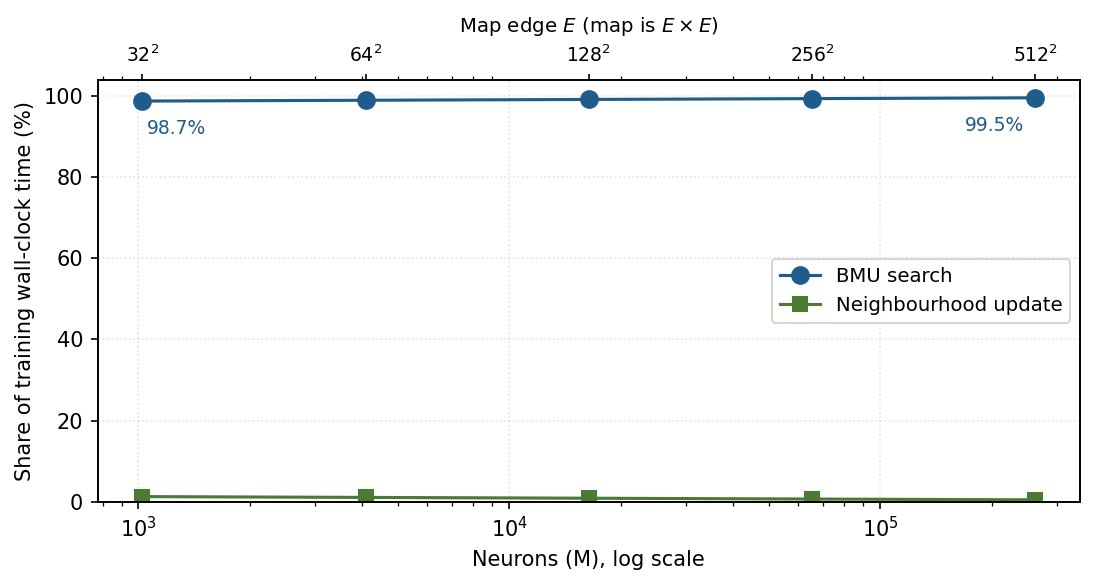}
\caption{Share of training wall-clock time spent in the BMU search versus the neighbourhood update, across the size sweep (log neuron axis). The BMU rises from 98.7\% of wall-clock time at 32$\times$32 to 99.5\% at 512$\times$512, while the factorised box-blur update never exceeds 1.3\%.}
\label{fig:3}
\end{figure}

\subsection{The efficiency chain, and the capacity wall}
To place the implementation against the alternatives on a strictly like-for-like footing, I fixed the work - 20 epochs over the full corpus, the same split, the same map - and measured wall-clock cost for every implementation that could run at all (Table~\ref{tab:6}, Figure 4). At 128$\times$128 the ordering spans nearly four orders of magnitude. The sparse-binary kernel needs 9.9 seconds per epoch. The sparse-float variant of the same design needs 27.0, so the binary specialisation is worth 2.1-2.8$\times$ across the ladder, though part of that is the halved FP16 codebook rather than binariness as such. The cuSPARSE feature-major baseline is competitive here at 9.3 seconds; its node-major counterpart needs 66.5. MedSOM, the CUDA implementation behind our earlier MEDLINE atlases, needs 792.8 seconds per epoch - 82$\times$ SparseBin.SOM - and somoclu, the best available multicore CPU library, needs 6,168.5 seconds on 28 threads, 621$\times$ SparseBin.SOM.

Three qualifications matter for reading this chain honestly. First, the advantage over the CPU library is not a fixed multiple but a trend that climbs steeply and then flattens: 85$\times$ at 32$\times$32, 349$\times$ at 64$\times$64, 621$\times$ at 128$\times$128 and 655$\times$ at 256$\times$256 - the last step adding only 5.5\%. Two effects compound. somoclu's per-epoch cost grows super-linearly in neurons - 6.2$\times$, 4.9$\times$ and 4.5$\times$ for successive quadruplings - while SparseBin.SOM grows sub-linearly at the smallest maps, where a 4090 is barely occupied, accelerating through 1.5$\times$, 2.8$\times$ and 4.3$\times$. The two rates converge near 4.3-4.5$\times$ by 256$\times$256, which is why the ratio plateaus at roughly 650$\times$ rather than continuing to climb. The gap widens with map size and settles near three orders of magnitude rather than starting there. I describe the GPU-versus-CPU advantage as approaching three orders of magnitude at the map sizes that matter, and give the whole curve rather than a single figure, since any one number is an artefact of the size at which it was taken. 

Second, the MedSOM figure is a timing comparison only: its per-epoch cost is measured on the same corpus and map, but I make no quality claim for it, since its destructive per-epoch normalisation renders its quantisation error non-comparable (2.2). 

Third, and most important for attribution, these are end-to-end system comparisons rather than layout results: MedSOM and somoclu differ from SparseBin.SOM simultaneously in codebook layout, in neighbourhood update - per-node loops against the factorised box blur - and in precision, both being FP32 by construction. 

The 82$\times$ and 621$\times$ figures therefore bundle all three, and should not be read as evidence for the layout argument; the isolated layout effect is the 4.5-8.5$\times$ of 5.1, measured within a single implementation with precision and update rule held fixed. Not every difference favours SparseBin.SOM: MedSOM's per-epoch normalisation makes its own search cheaper, so that difference works against the ratio rather than inflating it. 

The precision component can be bounded, though not cleanly removed, because the benefit of FP16 is itself layout-dependent (5.1). Matching downward, by running SparseBin.SOM at FP32, would forfeit its 1.65$\times$ and leave roughly 50$\times$; matching upward, by giving a node-major implementation FP16, would buy it little - 1.07$\times$ for the node-major cuSPARSE kernel - and leave roughly 77$\times$. A single precision-matched figure therefore does not exist; the ratio lies near 50-77$\times$, and the width of that interval is itself a consequence of the layout effect this paper argues for. The final row of the table is the one that most constrains practice - at 512$\times$512 both cuSPARSE layouts and MedSOM exhaust the 24 GB card as shipped, and SparseBin.SOM is the only one that runs without re-engineering. The barrier is not the score block, which is tunable to a few hundred megabytes, but two full-codebook FP32 update buffers on the device and a host-side initialisation that materialises the codebook twice before upload. Replacing both, the cuSPARSE path does train at 512$\times$512, in a comparable footprint and at roughly seven times the wall-clock cost at matched held-out quality (7).

\begin{table}[htbp]\centering
\caption{Per-epoch training cost at matched work.}
\label{tab:6}
\footnotesize
\begin{tabular}{lrrrrr}
\toprule
Implementation & 32$\times$32 & 64$\times$64 & 128$\times$128 & 256$\times$256 & 512$\times$512 \\
\midrule
SparseBin.SOM & 2.4 & 3.6 & 9.9 & 42.2 & 186.3 \\
cuSPARSE.SOM (feature-major) & 0.7 & 2.3 & 9.3 & 67.0 & OOM \\
SparseFloat.SOM & 4.9 & 8.6 & 27.0 & 112.9 & 480.4 \\
cuSPARSE.SOM (node-major) & 4.2 & 16.6 & 66.5 & 294.6 & OOM \\
MedSOM (prior CUDA SOM) & 37.6 & 197.8 & 792.8 & 3,209.9 & OOM \\
somoclu (CPU, 28 threads) & 202.9 & 1,251.5 & 6,168.5 & 27,685 & - \\
\bottomrule
\end{tabular}

\vspace{0.6ex}
{\footnotesize\raggedright Seconds per epoch at matched work (20 fixed epochs, full 26.9 million-article training split, RTX 4090; somoclu on 28 CPU threads at steady state). All GPU implementations are FP16 except MedSOM and somoclu, which are FP32 by construction. At 512$\times$512 only SparseBin.SOM runs; the others exhaust 24 GB. somoclu was measured on an uncontended run at all four sizes; when contention was tested at 64$\times$64 it cost under 2\%.\par}

\end{table}

\begin{figure}[htbp]\centering
\includegraphics[width=\linewidth]{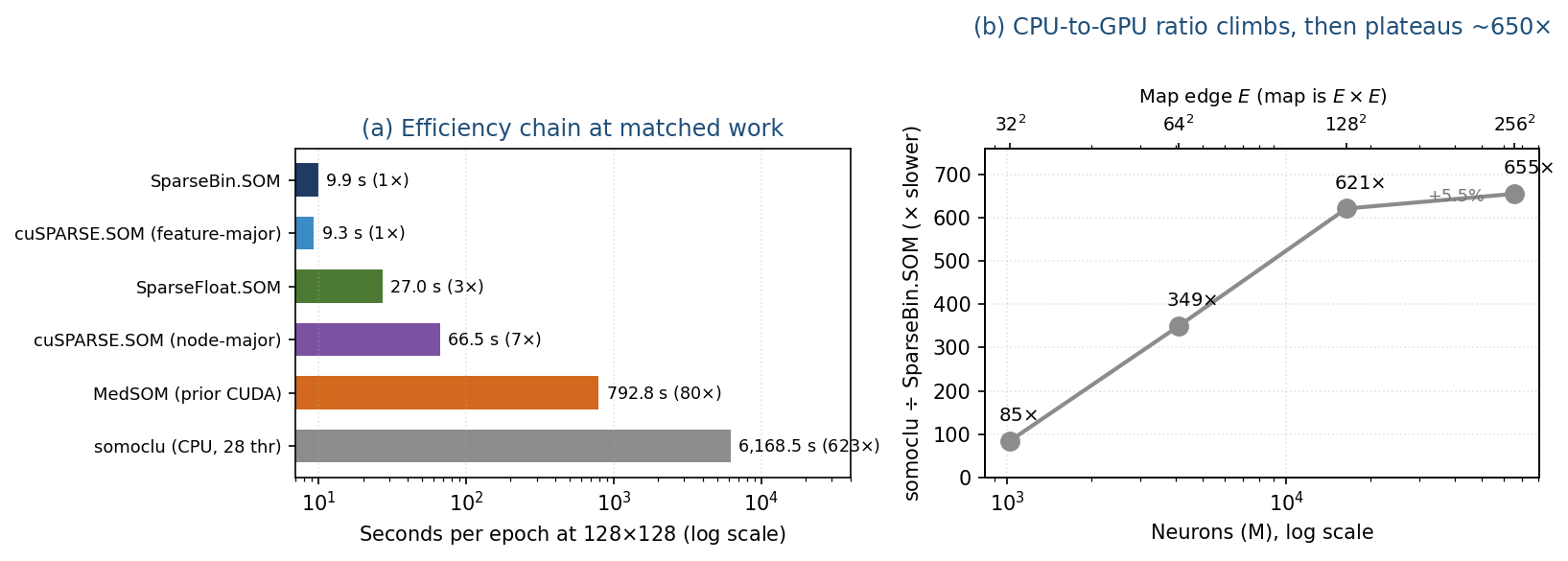}
\caption{(a) Per-epoch training cost at 128$\times$128 on the full corpus, matched at 20 epochs (log scale); the chain spans more than two and a half orders of magnitude from the best CPU library to the sparse-binary GPU kernel. (b) The CPU-to-GPU ratio is not a constant: it climbs steeply from 85$\times$ at 32$\times$32 to 621$\times$ at 128$\times$128, then flattens to 655$\times$ at 256$\times$256, because somoclu's super-linear scaling and its sub-linear scaling converge as the map grows.}
\label{fig:4}
\end{figure}

\subsection{Map-size scaling: a power law with no elbow}
The final experiment asks how quality scales with map size when nothing else varies. Every rung trains on the same full-corpus split with the same initialisation, precision, and stopping rule, doubling the lattice edge each time from 32$\times$32 to 512$\times$512 on the RTX 4090, and I add a 1,048,576-neuron map trained on an H200 (Table~\ref{tab:7}, Figure 5). Held-out quantisation error falls monotonically across the whole range, from 0.5241 to 0.3437, and the decline is a clean power law. Because the largest rung was trained under the earlier adaptive $\sigma$ controller rather than the deterministic schedule used for the rest, I report the fit both ways. Across the five deterministic rungs the exponent is $-$0.0597 (95\% CI $-$0.067 to $-$0.052) with R$^{2}$ = 0.996, a 4.05\% improvement in QE per doubling of the neuron count; including the H200 rung, which extends the range to three full decades, gives $-$0.0615 (95\% CI $-$0.067 to $-$0.056) with R$^{2}$ = 0.996, or 4.17\% per doubling. The two agree within each other's confidence intervals, so the extra point extends the curve without bending it.

The absence of an elbow is the substantive result, and because it is a negative claim I tested it rather than reading it off the plot. On the five deterministic rungs, fitting a segmented model with a free breakpoint \cite{muggeo2003} improves the residual sum of squares from 5.80$\times$10$^-$$^5$ to 9.81$\times$10$^-$$^6$. An F-test does not support that as a real improvement (F = 4.91, p = 0.27): the single power law is not rejected in favour of a two-regime fit at any interior knot. Adding the H200 rung the segmented model remains unsupported at the 5\% level, but the margin narrows (F = 15.5, p = 0.059, best knot at 128$\times$128), and I report this rather than the more convenient figure alone. Two observations settle how it should be read. 

First, the extra point does not create the curvature: the five deterministic rungs alone select the same breakpoint and almost the same pair of slopes ($-$0.0528 and $-$0.0666, against $-$0.0527 and $-$0.0667 with the sixth point), so the largest map adds statistical power to detect a departure that is already present rather than introducing one. 

Second, and decisively, the departure runs the wrong way to be an elbow. An elbow marks the point of diminishing returns - a shallower slope beyond the breakpoint, marking the size past which refinement stops paying. Here the slope steepens, from 3.59\% per doubling below the knot to 4.52\% above it: quantisation error improves faster as the map grows, not more slowly. Whatever mild two-regime structure the data contain therefore reinforces the conclusion rather than qualifying it, and no size in the range examined marks a point of diminishing return. Nor does the per-doubling gain show any sign of decaying toward a plateau: it runs 3.3\%, 4.0\%, 4.1\%, 4.8\%, 4.4\% across the five doublings, drifting up rather than down, and never approaches the 0.5\% threshold at which further refinement would cease to pay (Figure 5b). Two secondary measures confirm the maps stay healthy: topographic error rises gently from 0.010 to 0.045, and the dead-unit fraction from 9.4\% to 20.0\%, so four neurons in five remain active at the largest size.

\begin{table}[htbp]\centering
\caption{Held-out quantisation error across the size sweep.}
\label{tab:7}
\footnotesize
\begin{tabular}{lrrrrrrr}
\toprule
Edge & Neurons & Epochs & Wall (s) & QE (held) & QE (Eucl) & TE & Dead \% \\
\midrule
32 & 1,024 & 22 & 52.5 & 0.5241 & 2.8181 & 0.010 & 9.4 \\
64 & 4,096 & 20 & 71.5 & 0.4899 & 2.7592 & 0.018 & 8.7 \\
128 & 16,384 & 25 & 249.1 & 0.4518 & 2.6876 & 0.027 & 9.3 \\
256 & 65,536 & 25 & 1,062.5 & 0.4152 & 2.6105 & 0.033 & 10.7 \\
512 & 262,144 & 29 & 5,427.1 & 0.3764 & 2.5204 & 0.042 & 14.7 \\
1024* & 1,048,576 & 44 & 35,495.3 & 0.3437 & 2.4312 & 0.045 & 20.0 \\
\bottomrule
\end{tabular}

\vspace{0.6ex}
{\footnotesize\raggedright A single full-corpus split (26,912,934 training articles), PCA initialisation, Kaski-Lagus stop; all six runs converged. Edges 32-512 were trained on the RTX 4090 under the deterministic exponential schedule. *The 1024$\times$1024 rung was trained on an H200 SXM under the earlier adaptive $\sigma$ controller and is reported separately for that reason; it uses the identical corpus split, initialisation, precision, and stopping criterion, and differs only in the $\sigma$ decay path (see text). QE columns give held-out cosine QE and implementation-native Euclidean QE. Held-out QE falls monotonically across the whole range with no elbow.\par}

\end{table}

\begin{figure}[htbp]\centering
\includegraphics[width=\linewidth]{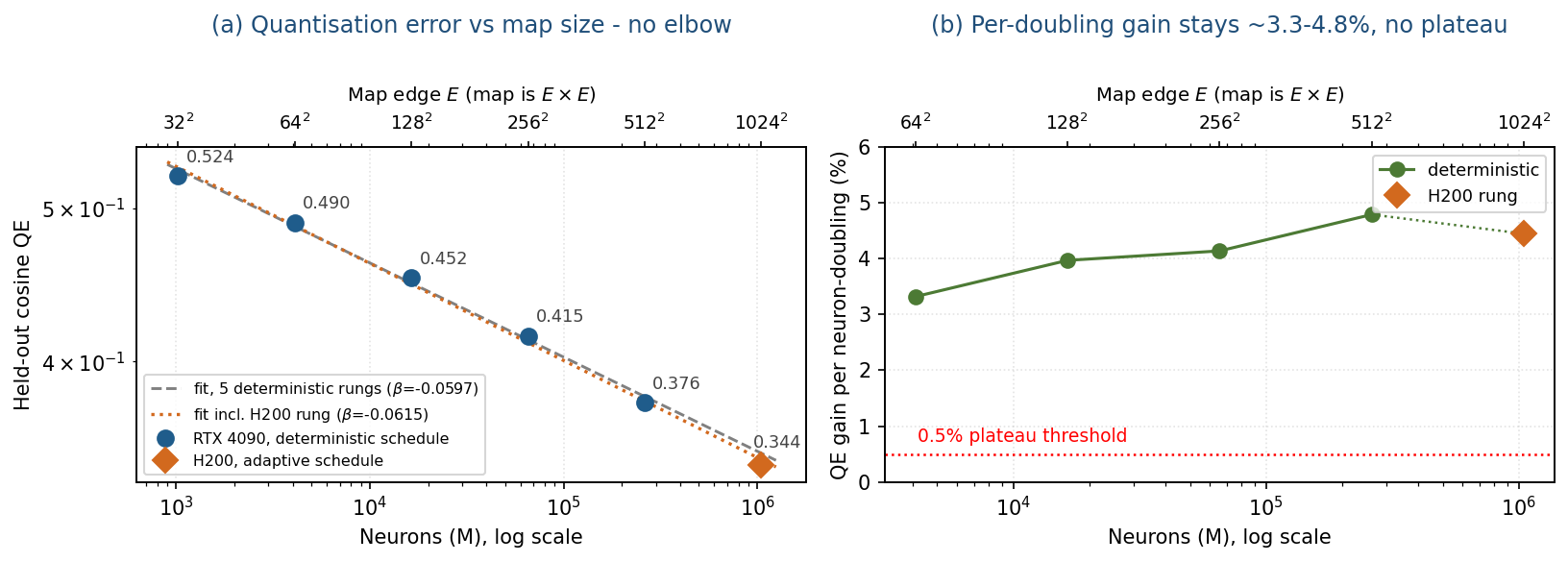}
\caption{(a) Held-out cosine QE against neuron count (log-log) from 32$\times$32 to 512$\times$512 on one full-corpus split, with the fitted power law. (b) QE improvement per neuron-doubling, drifting up from 3.3\% to 4.8\% and never approaching the 0.5\% plateau threshold.}
\label{fig:5}
\end{figure}

At 1,048,576 neurons the largest map exceeds the biggest self-organising map in the peer-reviewed record - the roughly 1.0 million-node WEBSOM patent map \cite{kohonen2000} - and does so over a corpus four times larger, 29.9 million articles against 6.8 million patent abstracts, and $\sim$30,000 item input vectors against $\sim$500. I am not aware of a larger trained SOM. The run was trained under the earlier adaptive $\sigma$ controller, so it is not schedule-matched to the rest of the ladder; and it is a single run, as are all rungs, so the sweep carries no interval. That is a weaker limitation than it sounds: PCA-initialised full-batch training on a fixed split is deterministic, so each rung is exact rather than a draw from a distribution, and re-running it reproduces the same value. 

\subsection{Where the advantage comes from, and where it does not}
Since the BMU search dominates training cost, I profiled the whole search phase with Nsight Compute \cite{nvidia_ncu} at two map sizes, summing every kernel each implementation runs - for the cuSPARSE path the codebook norm, the sparse-dense product and the argmin reduction; for SparseBin.SOM the single fused kernel that does all three (Table~\ref{tab:8}, Figure 6). The result overturns the mechanism I had previously proposed, and I report it as such. The natural explanation for a sparse-binary kernel's speed is that it simply moves fewer bytes: recording presence rather than values, and holding the codebook in half precision, ought to cut the DRAM traffic on which a bandwidth-bound search depends. An earlier version of this work reported exactly that, a 3.85$\times$ reduction in DRAM moved. The figure does not survive precision-matching. It compared an FP16 kernel against an FP32 baseline, so most of the gap was the precision difference rather than the representation, and with both implementations storing the codebook in half precision the traffic advantage disappears.

What replaces it is a crossover rather than a direction. At 128$\times$128 the fused kernel moves slightly less data than cuSPARSE feature-major - 2.82 TB against 3.10 TB per epoch, some 9\% less - while at 256$\times$256 it moves substantially more, 18.97 TB against 13.85 TB, or 37\% more. Sustained bandwidth crosses in the same place and the opposite way: at 128$\times$128 the cuSPARSE path streams faster (35.3\% of peak against 29.0\% here), whereas at 256$\times$256 SparseBin.SOM streams more than twice as fast (45.2\% against 21.5\%, with node-major managing 7.8\%). Neither traffic volume nor bandwidth efficiency is a fixed property of either design.

The per-kernel breakdown identifies the mechanism directly, and it is not the one I expected. Because the cuSPARSE path computes scores with a sparse-dense product, it must materialise the dense score block in memory and read it back to find each sample's best unit. That costs 0.86 TB written and 0.89 TB read at 128$\times$128, and 3.36 TB and 3.56 TB at 256$\times$256 - between a half and rather more than a half of everything that path moves. 

The fused kernel accumulates distances in registers and takes its two best units in the same sweep, writing no score block at all. The decisive observation is what that read-back costs in time. At 256$\times$256 the sparse-dense product streams at 67.7\% of Nsight Compute's measured sustained peak ($\approx$982 GB/s) and finishes 10.27 TB in 15.4 seconds; the reduction that follows it streams at 7.7\% of the same peak and takes 47.3 seconds to move 3.57 TB. The reduction is thus a quarter of the traffic and three-quarters of the time (Figure 6c). Reading a dense score block back is a poorly-localised access over a structure far larger than cache, and it is precisely this pass that fusion removes.

The advantage at large maps therefore comes from not materialising the score matrix rather than from moving less data, which vindicates the fusion rationale of 3.3 rather than the traffic argument I had built on top of it. The claim should be read as scoped to how that matrix is accessed: rewriting the argmin read-back so that it streams coalesced, and pairing it with a larger batch, removes most of the per-epoch advantage at 256$\times$256 (7). A separate instruction-level profiling addendum, using the same six captures with an extended counter set, locates the fused kernel more precisely still: it executes essentially no floating-point multiply, reaches only 30 to 46\% of the nearest ceiling at every level of the memory hierarchy and no more than 32\% of warp-issue peak, and is therefore latency-limited rather than bandwidth- or compute-bound, its 121 registers per thread holding occupancy to a third. That profiling requires a Nsight Compute-enabled image and elevated capabilities that the reproduction container deliberately does not carry, so it is released as a separate repository, \url{https://github.com/mongrolwarrior/sparsesom-roofline-addendum} at tag \texttt{v1.0}, rather than as part of the pipeline of 8; a written addendum accompanies this article as an ancillary file.

\begin{table}[htbp]\centering
\caption{Whole-BMU-phase profiling.}
\label{tab:8}
\footnotesize
\resizebox{\textwidth}{!}{%
\begin{tabular}{lrrrrr}
\toprule
Model & Edge & \shortstack{BMU\\(s/ep)} & \shortstack{DRAM/ep\\(TB)} & \shortstack{score\\block} & \shortstack{DRAM BW\\(\% of 1,008 GB/s)} \\
\midrule
SparseBin.SOM & 128 & 9.63 & 2.82 & - (fused) & 29.0 \\
cuSPARSE.SOM (feature-major) & 128 & 8.72 & 3.10 & 1.74 (56\%) & 35.3 \\
cuSPARSE.SOM (node-major) & 128 & 65.64 & 3.84 & 1.73 (45\%) & 5.8 \\
SparseBin.SOM & 256 & 41.61 & 18.97 & - (fused) & 45.2 \\
cuSPARSE.SOM (feature-major) & 256 & 63.80 & 13.85 & 6.92 (50\%) & 21.5 \\
cuSPARSE.SOM (node-major) & 256 & 289.30 & 22.86 & 7.02 (31\%) & 7.8 \\
\bottomrule
\end{tabular}}

\vspace{0.6ex}
{\footnotesize\raggedright Measured with Nsight Compute, full training split, RTX 4090, FP16 throughout. For cuSPARSE the phase sums the norm, sparse-dense product, row-pointer setup and argmin-reduction kernels; for SparseBin.SOM it is the single fused kernel plus a negligible norm pass. ``Score block'' is the dense score matrix the cuSPARSE path writes and reads back, which SparseBin.SOM never materialises. Sustained bandwidth here is phase DRAM divided by measured BMU time, expressed against the card's theoretical peak of 1,008 GB/s. Note that this is a different denominator from the per-kernel figures in Table~\ref{tab:9}, which are Nsight Compute's own dram\_\_throughput percentages against its measured sustained peak of $\approx$982 GB/s (97.4\% of theoretical); the two bases differ by 2.6\% and are not interchangeable. The 128$\times$128 captures are exact; at 256$\times$256 cuSPARSE.SOM (feature-major) is sampled at 1-in-22 launches and cuSPARSE.SOM (node-major) at 1-in-329, validated to within $\sim$7\% against the exact captures.\par}

\end{table}

\begin{table}[htbp]\centering
\caption{Per-kernel decomposition of the cuSPARSE BMU phase.}
\label{tab:9}
\footnotesize
\resizebox{\textwidth}{!}{%
\begin{tabular}{lrrrrr}
\toprule
Kernel & \shortstack{DRAM\\(TB)} & \shortstack{Share of\\traffic} & \shortstack{Time\\(s)} & \shortstack{Share of\\time} & \shortstack{DRAM BW\\(\% of $\approx$982 GB/s)} \\
\midrule
Sparse-dense product (writes score block) & 10.27 & 74\% & 15.44 & 25\% & 67.7 \\
Argmin reduction (reads score block back) & 3.57 & 26\% & 47.33 & 75\% & 7.7 \\
Norm + row-pointer setup & <0.01 & <1\% & 0.02 & <1\% & - \\
\bottomrule
\end{tabular}}

\vspace{0.6ex}
{\footnotesize\raggedright The mechanism in one table; all rows are the cuSPARSE feature-major path at 256$\times$256. Percentages here are Nsight Compute's per-kernel dram\_\_throughput against its measured sustained peak of $\approx$982 GB/s - a different denominator from Table~\ref{tab:8}, which uses the theoretical 1,008 GB/s; the two differ by 2.6\%. On that basis the sparse-dense product streams at 67.7\%; the argmin reduction that reads its score block back streams at 7.7\%, so a quarter of the traffic consumes three-quarters of the time. The fused kernel performs both roles in one pass at 45.6\% and never materialises the score block.\par}

\end{table}

\begin{figure}[htbp]\centering
\includegraphics[width=\linewidth]{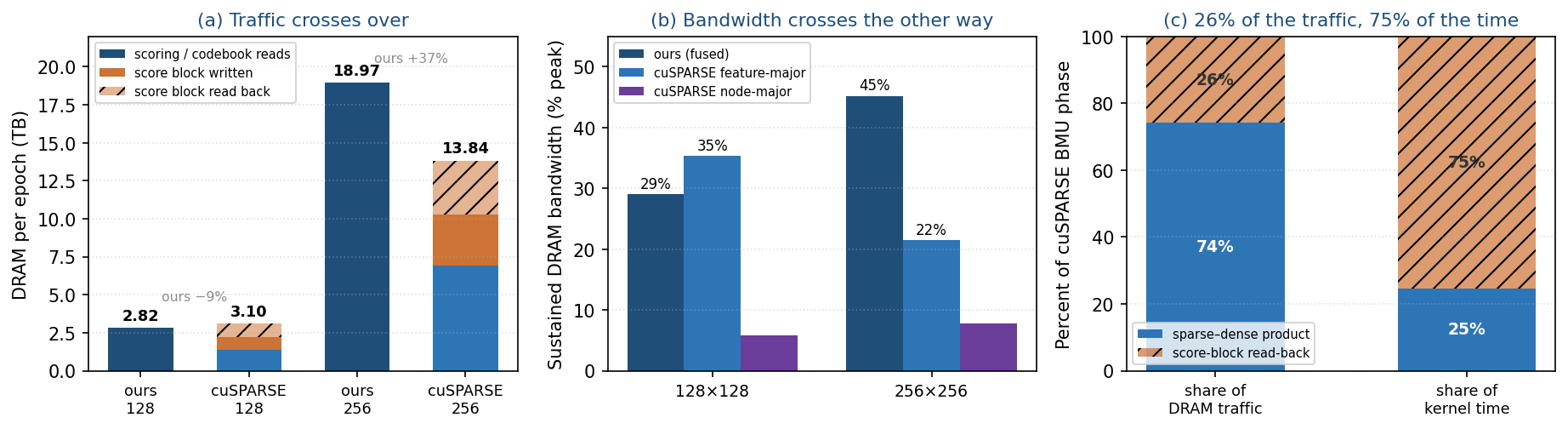}
\caption{Whole-phase BMU profiling, FP16-matched. (a) DRAM per epoch, with the cuSPARSE bars split to show the write-then-read-back of the dense score block the fused kernel never materialises: the present traffic is 9\% lower at 128$\times$128 and 37\% higher at 256$\times$256. (b) Sustained DRAM bandwidth, crossing over the other way. (c) The mechanism: within the cuSPARSE phase at 256$\times$256, the score-block read-back is a quarter of the traffic but three-quarters of the elapsed kernel time.}
\label{fig:6}
\end{figure}

\begin{figure}[htbp]\centering
\includegraphics[width=\linewidth]{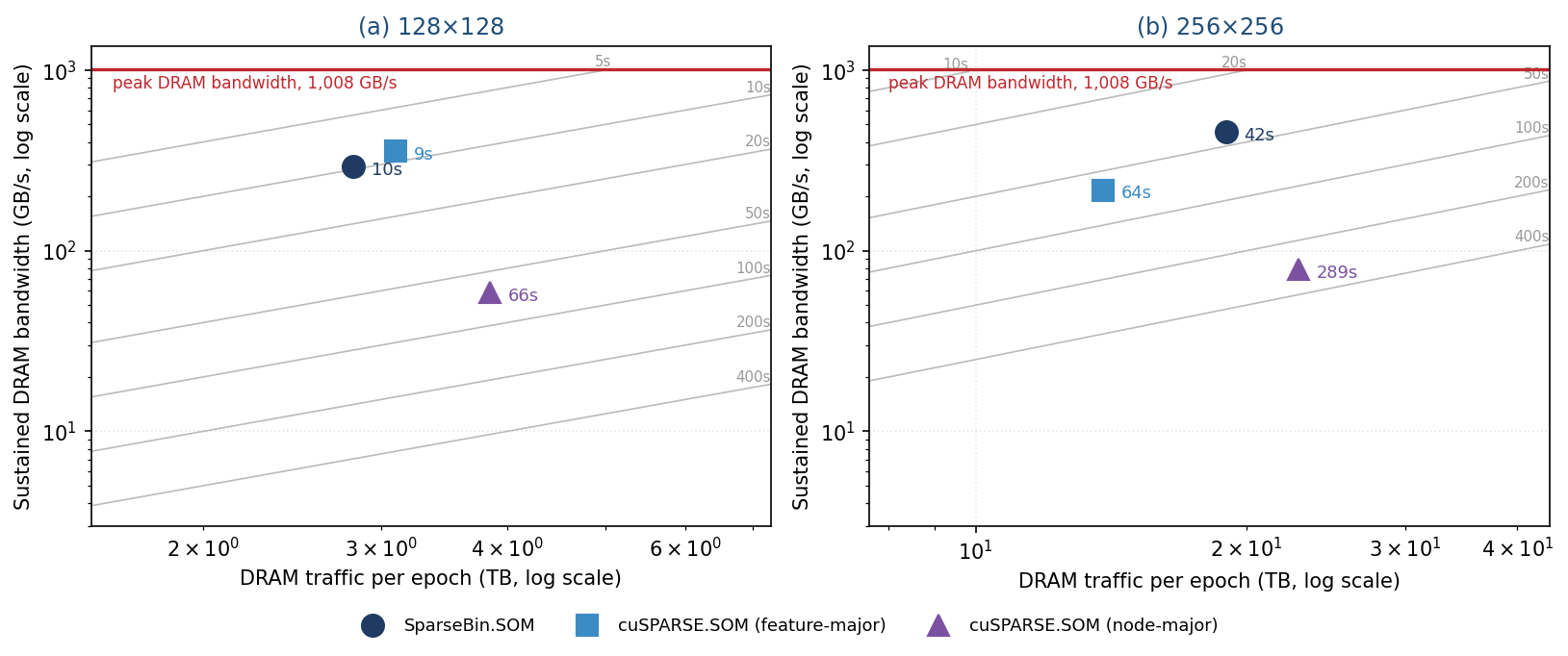}
\caption{DRAM traffic against sustained bandwidth for the whole BMU phase, at 128$\times$128 (a) and 256$\times$256 (b). The horizontal line is the theoretical peak of 1,008 GB/s; the grey diagonals are contours of equal elapsed time, and each marker is annotated with the measured phase time. Moving up is streaming faster; moving right is moving more bytes; moving toward the upper left is finishing sooner. At 128$\times$128 the three implementations separate almost entirely by bandwidth. At 256$\times$256 the fused kernel sits to the right of the cuSPARSE feature-major path, moving 37\% more data, yet well above it and on a faster iso-time contour, which is the crossover of 5.7 in one view.}
\label{fig:7}
\end{figure}

\section{Discussion}
Taken together, the results cohere around a single mechanism, though not the one I first proposed. Training a self-organising map at this scale is almost entirely a problem of moving the codebook through memory: the best-matching-unit search reads it on every epoch, and at a codebook far larger than the cache it is the access pattern, not the arithmetic, that decides the cost. Organising the codebook feature-major addresses exactly this, turning the search into a coalesced, tiled product in which each loaded weight column is reused across the samples of a tile - worth 4.5-8.5$\times$ on the search when nothing else is varied (5.1). Because the best-matching unit is an exact argmin, none of this touches quality; the speed-up is, in the strict sense, free. Quality is instead governed by the update, which I keep inexpensive with a radius-independent box blur.

What the measurements then revise is why the complete design wins at large maps. The intuitive account - that a compact binary representation simply moves fewer bytes - does not survive precision-matching. At 256$\times$256 SparseBin.SOM moves 18.97 TB of DRAM per epoch against the cuSPARSE baseline's 13.85 TB, some 37\% more, and is nonetheless faster: 41.6 seconds against 63.8. Nor is the difference in the gaps between kernels, which whole-phase profiling bounds at a few per cent. It lies in what kind of traffic each design is obliged to move.

A sparse-dense product must materialise the dense score block in memory and read it back to find each sample's best unit, and that round trip alone costs 6.92 TB - half of everything that path moves - and, because the read-back is poorly localised, three-quarters of the elapsed kernel time. A fused search accumulates in registers and takes its two best units in the same sweep, so it pays nothing for that class of traffic at all, and streams the traffic it does move at more than twice the sustained bandwidth (45.2\% against 21.5\% of peak). The advantage is therefore the elimination of a category of memory traffic, not a reduction in the total; the absolute volume here is higher, and saying so is what makes the mechanism unambiguous.

Read this way the two design choices of Section 3 attack the same bottleneck by different means. Feature-major storage does not reduce the bytes the codebook read requires: it lets a warp fetch them in a few wide, cache-line-aligned transactions instead of many narrow ones, so the same traffic is delivered at a far higher fraction of peak bandwidth. Fusing the score and the argmin into a single pass does reduce the bytes, because the scores never reach memory at all and a whole class of traffic is never generated. One choice improves how efficiently data moves; the other removes data that would otherwise have to move. It is the combination - how the weights are laid out, and what is never written down - rather than any single clever kernel, that makes a full-corpus MEDLINE atlas tractable on commodity hardware. That combination carries it past a quarter of a million neurons on one consumer card, and beyond a million on a single datacentre GPU.

The approach is not specific to binary data. Replacing the gather-sum of the binary inner product with a gather-multiply - carrying $\|$x$\|$$^{2}$ and a values array alongside the feature indices - would extend the same feature-major, fused-top-two, factorised-update design to sparse real-valued inputs at little additional cost in the large-map regime, where the shared O(M) codebook read already dominates. The binary specialisation is most valuable at small and mid map sizes; a sparse-float variant would trade a little of that advantage for considerably broader applicability, which I regard as a worthwhile exchange and a natural next step.

A more conceptual implication concerns how the size of such a map ought to be chosen. The absence of an elbow in quantisation error is easily mistaken for an embarrassment - a failure to find the ``right'' size - when it is in fact a substantive finding. On a corpus whose topical structure is heavy-tailed and self-similar rather than concentrated at one granularity \cite{clauset2009}, there is no characteristic granularity for a flat grid to discover, so representational error simply improves with resolution until the compute budget is exhausted. The size question becomes meaningful only once a downstream objective is specified, and here the two objectives pull apart. Faithful quantisation rewards arbitrarily large maps; faithful recovery of the coarse knowledge structure, which I examine in the companion paper \cite{amosprep}, saturates on quite small maps, because a flat lattice cannot embed an exponentially branching hierarchy beyond a shallow depth.

A concrete case makes the dissociation tangible. With roughly 29.9 million articles, a 50$\times$50 map holds on average some twelve thousand papers per neuron, a 512$\times$512 map about a hundred, and the 1,048,576-neuron map reported here roughly twenty-eight. 

Consider the literature on CAR-T cell immunotherapy: even a 50$\times$50 map places it correctly beside the rest of cancer immunotherapy and far from, say, cardiology, which is entirely adequate for browsing. At that resolution, however, the whole of it - B-cell-lymphoma trials, myeloma trials, bispecific T-cell engagers, the management of cytokine-release syndrome - collapses onto one or two neurons whose prototype is an average over twelve thousand papers, and these distinct research fronts become indistinguishable. Enlarging the map gives each front its own contiguous patch of neurons, so that near-duplicate detection, fine-grained nearest-neighbour retrieval, and the emergence of a genuinely new front on a previously empty neuron all become possible at the granularity of the front rather than the subfield. Recovering that the CAR-T region sits where it belongs is a small-map property; resolving within it is a large-map one, and it is this within-region resolution that quantisation error keeps rewarding without an elbow.

Finally, it is worth returning to the purpose that motivates the work. The value of a self-organising map of the medical literature was never its quantisation error in the abstract, but its promise as an objective, browsable atlas of the kind that science mapping has long sought \cite{borner2010}. Against such an atlas, expert judgement about what to teach, to fund, and to research can be checked and, where necessary, corrected. That promise has been constrained less by theory than by the practical difficulty of building such a map over the whole corpus, at a resolution fine enough to be useful. The method described here removes much of that constraint: a full-MEDLINE atlas is now a matter of minutes on a single GPU, and its resolution can be set by the available compute rather than by a data-imposed ceiling. Whether the resulting atlas faithfully encodes the biomedical knowledge structure - against the MeSH tree and against independent, citation-derived taxonomies - is the question taken up in the companion validity paper \cite{amosprep}.

\section{Limitations}
Several limitations qualify these results. All measurements were made on a single consumer architecture, the RTX 4090; the layout argument should hold on datacentre GPUs but I have not confirmed it there, and cross-GPU BMU identity is argued from the exactness of the argmin rather than demonstrated. The per-epoch advantage over cuSPARSE appears only above roughly 128$\times$128 at the configurations compared here - below that the baseline is faster, and the wall-clock advantage at 128$\times$128 rests partly on converging in fewer epochs rather than on raw speed. With the tile size retuned the boundary moves down between 32$\times$32 and 64$\times$64 (7). Epoch counts themselves carry two sensitivities that are not stated with the tables: changes to floating-point reduction order that leave quality unaltered move the converged count by about five epochs, and the randomly initialised baseline varies by up to about 1.4$\times$ between extreme seeds while its held-out error varies by under 0.2\%. The DRAM comparison crosses over between the two profiled map sizes, so neither implementation can be described as uniformly leaner, and the 256$\times$256 traffic figures are sampled and scaled rather than captured exhaustively (validated to within $\sim$7\% against the exact 128$\times$128 capture). Profiling covers two map sizes on one architecture; whether the same crossover appears at other sizes or on datacentre GPUs is untested. The somoclu comparison rests on a single map size, chosen as the most favourable point for it. The MedSOM comparison is timing-only, since its per-epoch codebook normalisation makes its quantisation error non-comparable. The size sweep is a single seed per rung, which is exact rather than uncertain because PCA-initialised full-batch training is deterministic, but it means the sweep carries no interval. Finally, kernel-shape, GPU-versus-CPU, and initialisation asymmetries between implementations are disclosed rather than eliminated.

\section{Since submission: tuning both implementations}
The measurements above compare two implementations at the configurations each was running in. After submission I ran a symmetric tuning programme: every optimisation lever proven on one side was offered to the other, both implementations were retrained to convergence, and the fused kernel was reprofiled. A companion addendum reports it in full, together with a corrigendum listing the statements above that it narrows (8).

Four levers proved to matter for SparseBin.SOM, none of them layout: the sample-tile size (the published choice of 16 is optimal at no map size tested), clustering the corpus so that a tile's feature union shrinks, splitting the neuron axis into chunks so the co-resident working set fits L2, and vectorised half-precision loads. Together they accelerate the search by 5.6 to 10.1$\times$ per epoch relative to the configuration benchmarked here. In whole-run terms, the converged 64$\times$64 map of Table~\ref{tab:7} falls from 71.5 s to 12.8 s at the same twenty epochs, a mean over three train/held-out splits. The comparisons of 5.5 move with it. MedSOM and somoclu are unchanged frozen binaries measured on the same machine, corpus and split under the same fixed-work protocol, so their ratios follow directly: against MedSOM the margin at 128$\times$128 rises from 82$\times$ to roughly 385$\times$, and against somoclu the curve of 5.5 becomes 655$\times$, 2,235$\times$, 2,994$\times$ and 3,643$\times$ at 32$\times$32 through 256$\times$256, so the CPU comparison moves from approaching three orders of magnitude to sitting between three and four. On the baseline's side the largest single gain was not in cuSPARSE at all but in the argmin read-back kernel written for this work, which assigned one thread per sample row and was therefore uncoalesced in exactly the manner 3.2 identifies as the central error; rewritten, it runs 12.5$\times$ faster at 256$\times$256, and the tuned baseline is 1.9 to 3.2$\times$ faster than the configuration compared here.

Against that tuned baseline the crossover of Table~\ref{tab:cross} does not merely widen, it disappears: SparseBin.SOM is faster at every map size, by 1.27$\times$ at 32$\times$32 to 4.22$\times$ at 256$\times$256 on converged wall clock at matched held-out quality, and by about 7$\times$ at 512$\times$512 against a baseline re-engineered to fit there. Those ratios are against a tuned opponent and are not comparable with the ratios in Table~\ref{tab:cross}, which are against its published configuration.

The programme also bounds what remains. The tuned kernel is the first configuration in this line of work to press a hardware ceiling rather than sit beneath every one: it reaches 77\% of measured L2 bandwidth, with warp-issue at 64\% and DRAM at 41\%, which limits any further lever of this kind to roughly 1.3$\times$ on this device. Every optimum reported here is tuned against the RTX 4090's cache and register file; the directions should transfer to other architectures and the values should not.

\section{Data and code availability}
The anonymised corpus is available from Zenodo at \url{https://doi.org/10.5281/zenodo.20770707} under a CC0 1.0 public-domain dedication. This is the permanent version of the DOI that resolves to the current version of the record. The corpus is distributed unsplit, in the sparse-binary CSR format described in 4.2, and contains only binary MeSH descriptor vectors: no PubMed identifiers, no titles, no abstracts and no author or journal data. The 90/10 partition used throughout 5 is regenerated deterministically from it by the reproduction pipeline at seed 42, which also verifies the resulting file hashes against those recorded in the frozen manifest. The Zenodo deposit carries the corpus only. The frozen result files behind every table in 5 are released in that repository under \texttt{frozen/2026-08-02/}, together with the SHA-256 manifest that hashes them, so each reported figure can be traced to the file that produced it. Appendix~\ref{app:provenance} gives the mapping.

The corpus is derived from the PubMed 2026 annual baseline, courtesy of the U.S. National Library of Medicine. It reflects that baseline and does not reflect the most current data available from NLM; the U.S. National Library of Medicine has not endorsed this work, and any errors of derivation are the author's.

The implementation, the cuSPARSE baseline, the shared evaluator and the reproduction pipeline are released under the MIT licence at \url{https://github.com/mongrolwarrior/sparsesom-paper1} at tag \texttt{v1.1} (DOI \texttt{10.5281/zenodo.22245649}), with pinned submodule commits recorded in the run provenance. Tag \texttt{v1.0} is the state that produced the frozen results and is what the first version of this article cites; \texttt{v1.1} differs from it only in the verification layer, rewording the claim set to match the present text and adding a manifest that records which claim is checked in which repository. No pipeline code and no data differ between the two, which is why the reworded claims verify unchanged against the same frozen results.

The verifier reports its two surfaces separately rather than as one total: thirteen claims are checked locally against those results, and three - the tuned per-edge comparison, the converged wall-clock table, and the stopping-rule substitution of 7 - are marked delegated, each naming the repository, tag and script that checks it. Those three are verified in the tuning addendum's repository, \url{https://github.com/mongrolwarrior/sparsesom-tuning} at tag \texttt{v1.0} (DOI \texttt{10.5281/zenodo.22245713}), which also carries both companion documents, the per-phase measurements behind them, and a re-derivation script that rebuilds the addendum's two headline tables from a fresh clone. The instruction-level profiling of 5.7 is at \url{https://github.com/mongrolwarrior/sparsesom-roofline-addendum} at tag \texttt{v1.0} (DOI \texttt{10.5281/zenodo.22245687}).

Both repositories were verified as a reader would encounter them, not as the author holds them: a credential-free clone of the public URL, a container build from the committed scripts alone, and the corpus fetched from the concept DOI above. For the pipeline at \texttt{v1.1} that exercise returned seventeen checks passing, none failing, and one skipping by design, the roofline check requiring a profiler the reproduction image deliberately does not carry. 

The pipeline provides two reproduction depths. The \texttt{outline} profile runs a single seed at reduced epoch budgets on the cheapest map size at which each effect is visible, and takes about three hours on one 24 GB GPU; it checks the direction and order of magnitude of every reported effect and reproduces no published value, range, interval or test. The \texttt{full} profile runs the complete matrix - five seeds, edges 32 to 512, every implementation - and takes approximately 6.5 days on the same card, after which the verification step scores the published claims; it may be run piecewise as four non-overlapping sections rather than in one pass. Nothing short of the full profile reproduces a statistic. Roofline analyses in the conventional arithmetic-intensity form are not included here, because the profiling captures behind 5.7 recorded memory traffic and timing but not floating-point counts; they are provided in the roofline addendum cited above. 

\section{Declarations}
Author contributions. A.J.A. is the sole author and is responsible for the conception and design of the work, the implementation, the experiments, the analysis, and the writing.

Acknowledgements. The MEDLINE self-organising map programme on which this work builds was developed with Kyungmi Lee, Tarun Sen Gupta and Bunmi S. Malau-Aduli, whose contributions to the earlier studies cited here \cite{amos2021,amos2022,amos2024a,amos2024b} shaped the questions this paper addresses. The large-map experiments used a rented NVIDIA H200 instance. This research is indexed against data courtesy of the U.S. National Library of Medicine.

Funding. This research received no specific grant from any funding agency in the public, commercial or not-for-profit sectors.

Competing interests. The author declares no competing interests.

Ethics. This study analysed publicly available bibliographic metadata (MeSH descriptor annotations from the PubMed 2026 baseline). It involved no human participants, no animal subjects and no identifiable personal data, and required no ethics approval.

Declaration of generative AI in the manuscript preparation process. During the preparation of this work the author used generative AI assistance (Anthropic Claude) to draft and revise manuscript text, to structure and restructure sections, and to verify and format reference metadata. After using these tools the author reviewed and edited the content as needed and takes full responsibility for the content of the article. Use of the same assistance within the research process itself is described in 4.1. No AI system is credited as an author, and none is accountable for the work.

\appendix
\section{Provenance of tables and figures}
\label{app:provenance}
Every table and figure in this paper is computed from a file in the frozen result set described in
Section 8. The mapping is given below so that any quantity can be traced to the file it
came from and recomputed by the scripts in the release.

\begin{table}[htbp]\centering
\small
\begin{tabular}{@{}>{\raggedright\arraybackslash}p{0.13\linewidth}>{\raggedright\arraybackslash}p{0.36\linewidth}>{\raggedright\arraybackslash}p{0.42\linewidth}@{}}
\toprule
Item & Content & Source file \\
\midrule
Table~\ref{tab:1} & Corpus statistics & \texttt{corpus\_manifest.json} \\
Table~\ref{tab:2} & Codebook layout in isolation & \texttt{impl\_compare.parquet} \\
Table~\ref{tab:cross} & Equal-quality crossover & \texttt{impl\_compare.parquet} \\
Table~\ref{tab:3} & Equal-quality comparison & \texttt{impl\_compare.parquet}, \texttt{efficiency\_sweep.parquet} \\
Table~\ref{tab:4} & Effect of the update rule & \texttt{impl\_compare.parquet} \\
Table~\ref{tab:5} & Neighbourhood-update cost & \texttt{impl\_compare.parquet} \\
Table~\ref{tab:6} & Per-epoch cost at matched work & \texttt{efficiency\_sweep.parquet}; somoclu clean re-measurement \\
Table~\ref{tab:7} & Held-out QE size sweep & \texttt{size\_sweep.parquet}; H200 run log \\
Table~\ref{tab:8} & Whole-BMU-phase profiling & \texttt{roofline\_phase\_summary.csv} \\
Table~\ref{tab:9} & Per-kernel decomposition & \texttt{roofline\_phase\_breakdown.csv} \\
Figure~\ref{fig:3} & BMU share of training time & \texttt{size\_sweep\_epochs.parquet} \\
Figure~\ref{fig:4} & Efficiency chain & \texttt{efficiency\_sweep.parquet} \\
Figure~\ref{fig:5} & Map-size sweep & \texttt{size\_sweep.parquet} \\
Figure~\ref{fig:6} & Roofline profiling & \texttt{roofline\_phase\_summary.csv}, \texttt{roofline\_phase\_breakdown.csv} \\
\bottomrule
\end{tabular}
\end{table}

Figures~\ref{fig:1} and~\ref{fig:2} are schematics and carry no measured data. Two further files in the same set support claims made in prose rather than in a table: \texttt{sigma0\_sweep.parquet} for the $\sigma_0$ and initialisation comparison of 3.4, and \texttt{bringup\_gate.json} for the bit-exactness check of 5.1.

\end{document}